\documentclass[sigconf,natbib=true,anonymous=false]{acmart}

\renewcommand\footnotetextcopyrightpermission[1]{} 
\usepackage{graphicx}
\usepackage{balance}  
\usepackage{xspace}
\usepackage{xcolor}
\usepackage{alltt}
\usepackage{listings}
\usepackage{color}
\usepackage{caption}
\usepackage{subcaption}
\usepackage{tabularx, tabulary}
\usepackage{makecell}
\usepackage{booktabs}
\usepackage[noend]{algpseudocode}
\usepackage{algorithm}
\usepackage{multirow}
\usepackage{balance}
\usepackage{adjustbox}
\usepackage{wrapfig}
\usepackage{enumitem}

\newcommand{\heading}[1]{\vspace{4pt}\noindent\textbf{#1}}

\makeatletter
\def\@copyrightspace{\relax}
\makeatother

\definecolor{zqzbackground}{rgb}{1,1,1}
\definecolor{zqzbasic}{rgb}{0.1,0.1,0.1}
\definecolor{zqzcomment}{rgb}{0.5,0.5,0.5}
\definecolor{zqzapi}{rgb}{0,0.5,0}
\definecolor{zqzstring}{rgb}{0.5,0,0}
\begin{document}

\setlength{\abovedisplayskip}{3pt}
\setlength{\belowdisplayskip}{3pt}

\newcommand{\name}[0]{{\fontfamily{bch}\selectfont DARQ}\xspace}

\title{An Empirical Study of the Impact of Federated Learning on Machine Learning Model Accuracy}

\author{Haotian Yang, Zhuoran Wang, Benson Chou, Sophie Xu, Hao Wang$^\S$, Jingxian Wang$^\dagger$, Qizhen Zhang\\{\normalsize University of Toronto, $^\S$Stevens Institute of Technology, $^\dagger$National University of Singapore}}

\begin{abstract}

Federated Learning (FL) enables distributed ML model training on private user data at the global scale.
Despite the potential of FL demonstrated in many domains, an in-depth view of its impact on model accuracy remains unclear.
In this paper, we investigate, systematically, how this learning paradigm can affect the accuracy of state-of-the-art ML models for a variety of ML tasks.
We present an empirical study that involves various data types: text, image, audio, and video, and FL configuration knobs: data distribution, FL scale, client sampling, and local and global computations.
Our experiments are conducted in a unified FL framework to achieve high fidelity, with substantial human efforts and resource investments.
Based on the results, we perform a quantitative analysis of the impact of FL, 
and highlight challenging scenarios where applying FL degrades the accuracy of the model drastically and identify cases where the impact is negligible.
The detailed and extensive findings can benefit practical deployments and future development of FL.
\end{abstract}

\maketitle

\section{Introduction}
\label{sec:intro}

Most of today's data is generated by Internet edges and endpoint devices such as smart phones, 
cameras, IoT sensors, and autonomous vehicles~\cite{idc_report}.
Harnessing data at the global scale is challenging due to the Internet and data centers' capacity constraints, and users' privacy concerns---uploading all of the world's data to the cloud for processing is unrealistic.
Federated Learning (FL)~\cite{fl_origin} has emerged as an effective solution to train machine learning (ML) models using global data.
In this paradigm, a centralized server initializes and distributes a model to clients.
The server and the clients then coordinate to train the model iteratively.
In each round, a subset of clients are selected to perform local training, potentially for multiple epochs, to update the model parameters with their private data.
At the end of each round, every participating client sends its new parameters to the server to update the global model, e.g., via aggregation.
The new global model is broadcast to clients for the next training round, repeating until a termination condition is met.
FL eliminates the need for transferring data to the cloud, thereby enhancing system efficiency and privacy preservation while facilitating scalable applications of ML.
The utility of FL has been explored in many domains~\cite{fl_app_med1, fl_app_rec1, fl_app_rec2, fl_app_rec3, fl_app_rec4, fl_app_rec5, fl_app_rec6, fl_app_smartcity1, fl_app_smartcity2, fl_app_fin1, fl_app_fin2, fl_app_edge1, fedspace, fl_app_llm1, fl_app_llm2}.

Despite the promises of FL, 
imbalanced data, e.g., non-IID label distribution and skewed data sizes across clients, and heterogeneous client environments, e.g., diverse computing and network capabilities, 
pose unique challenges.
A vast amount of work has been dedicated to advance various aspects of FL, including communication efficiency~\cite{fl_opt_comm1, fl_opt_comm2, fl_opt_comm3}, fairness~\cite{fl_opt_fair1, fl_opt_fair2, fl_opt_fair3, fl_opt_fair4}, and privacy~\cite{fl_opt_priv1, fl_opt_priv2, fl_opt_priv3, fl_opt_priv4, fl_opt_priv5, fl_opt_priv6}.
The question we seek to answer in this paper is: {\em how does this learning paradigm for the global scale affect the accuracy of ML models that have shown high accuracy in centralized learning?}
Indeed, privacy-preserving, decentralized training in FL degrades model accuracy.
Even when provisioned with the same architecture and initialization, a model trained with FL is less accurate than one trained with centralized learning~\cite{fl_origin}.
While the answer to the above question is crucial to practical adoption, and the overhead of FL on model accuracy has been reported in previous work for specific ML tasks~\cite{fl_acc1, fl_acc2, fl_acc3, fl_acc4, fl_acc5, fl_acc6, fl_acc7, fl_acc8, fedadam}, there is a lack of a comprehensive evaluation of the impact of FL on the accuracy of {\em state-of-the-art models} for {\em different ML tasks}.

To fill this gap, this paper presents an empirical investigation on model accuracy degradation due to FL.
The distinction of this study is reflected in the following aspects.
(1) ML workloads: we first cover all common types of sensitive data from mobile and IoT applications--text (e.g., messages), image (e.g., photos), audio (e.g., recordings), and video (e.g., vlogs)--selecting a representative ML training task for each.
(2) ML models: for each ML task, we select a state-of-the-art model that achieves the highest accuracy from public leaderboards.
(3) FL platform: we adopt a unified, portable and production-ready FL framework to run experiments with high fidelity and adapt our selected models for it. 
(4) FL configuration: based on the overall architecture, we investigate a range of configuration knobs that may affect the FL-trained model accuracy.
(5) Resource investment: in total the experiments in this study consumed around 
6.2\,K GPU hours.


More specifically, we perform quantitative analysis of the impact of FL on ML model accuracy in five dimensions.
The first and perhaps the most critical characteristic of FL for accuracy is unbalanced data between clients: different clients generate different data (i.e., non-IID data distribution) and different amounts of data (i.e., skewed data volumes).
This is in contrast to centralized learning, where a training job accesses the entire data set.
Our results validate the significance of non-IID data distribution and show that the impact of volume skewness is limited.

We follow the FL workflow to investigate client sampling next, which trades between training speed and learning efficiency and also plays a role in model accuracy.
Conceptually, with more FL clients sampled in a training round, the training data can be more likely to represent global distribution, and thus the trained model is more accurate.
With a proper rate and strategy, however, sampling may not necessarily compromise model accuracy~\cite{fl_origin, oort}.
We also find that as long as a significant portion of client being selected, sampling clients incurs virtually no cost on accuracy.

We further investigate the scale of FL, the number of participating clients.
Intuitively, the larger the degree of distribution, the less effective FL can maintain model accuracy due to lower learning efficiency at each client.
However, in our experiments, we find that while scale affects FL accuracy in the short term, the accuracy of models trained with enough rounds eventually catches up.

Client-local learning strategies like batch size and training length affect how well the model learns client data. While we found these are critical hyperparameters for centralized learning, their impacts diversify on FL's global model. 
Finally, we evaluate the efficacy of different global federation approaches.
In particular, we compare the original FedAvg to two more recent and optimized FL algorithms: FedAdam~\cite{fedadam}, which incorporates adaptive learning rates using first- and second-moment estimates of gradients, and FedYogi~\cite{fedadam}, which builds on FedAdam by controlling variance via bounded updates.
Perhaps surprisingly, we find that what FedAvg achieves is on par with the more recent proposals and can be even better. 


In summary, this paper presents a detailed study of how accurate state-of-the-art models for various tasks may behave when adopted for FL, and what FL configuration knobs can have major impacts.
We make specific contributions as follows.

\begin{itemize}[noitemsep,topsep=0pt,parsep=0pt,partopsep=0pt]
    \item Workload, model, and framework selection for a holistic investigation of model accuracy in FL.
    \item Implementing and open-sourcing the FL version of state-of-the-art ML models for texts, images, audios, and videos.
    \item Extensive evaluation of the impact of FL on ML model accuracy, which required 6.2\,K GPU hours.
    \item Thorough result analysis, which sheds light on the implications of various FL knobs for popular ML tasks.
\end{itemize}

\section{Background}
\label{sec:background}

\begin{figure}
    \centering
    \includegraphics[width=0.85\columnwidth]{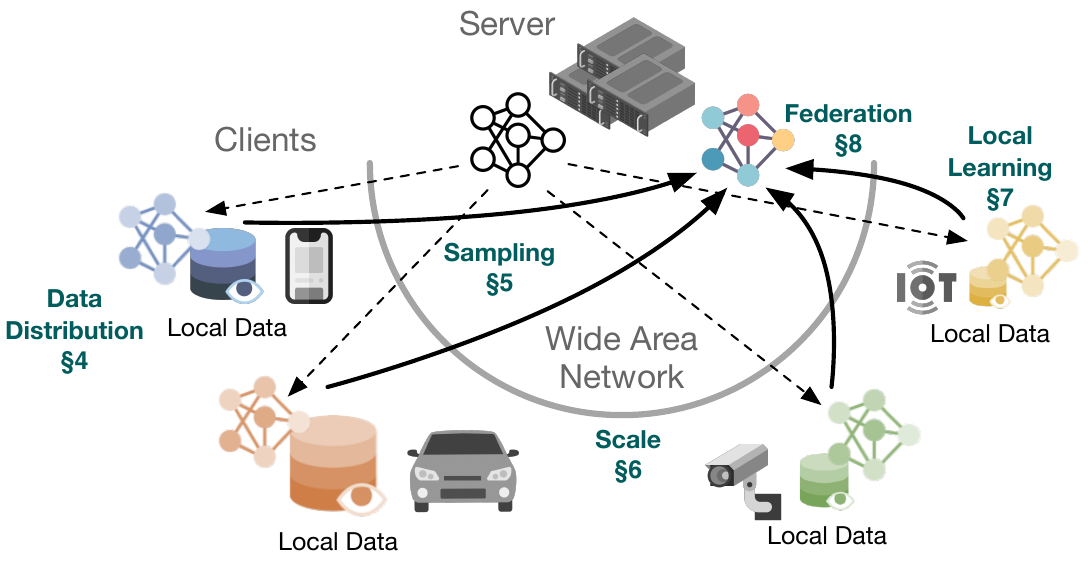}
    \caption{FL overview. Our investigation starts with data distribution and then follows the order of execution workflow.}
    \label{fig:fl_arch}
\end{figure}

Figure~\ref{fig:fl_arch} shows the overall architecture of FL, where a FL server communicates with a set of heterogeneous edge devices 
loosely-connected via a wide-area network. 
The goal of FL is to solve a joint optimization problem as
\begin{align}\label{eq:opt-fl}
\min_{\boldsymbol{w}\in\mathbb{R}^d} \mathcal{L}(\boldsymbol{w}, \mathcal{D}):= \sum_{j\in\mathcal{N}} \frac{|\mathcal{D}_j|}{|\mathcal{D}|} \mathcal{L}_j(\boldsymbol{w},\mathcal{D}_j),
\end{align}
where $\boldsymbol{w}$ denotes the model parameters, $\mathcal{N}$ denotes the set of clients, $\mathcal{D}_j$ is the local dataset of client $j\in\mathcal{N}$, the entire training dataset is $\mathcal{D}=\cup_{j\in\mathcal{N}}\mathcal{D}_j$, and $\mathcal{L}_j(\boldsymbol{w},\mathcal{D}_j)$ is the local loss function of client $j$. 
%

To solve this problem, the FL server first initializes a random global model $\boldsymbol{w}_0$, then oversees the training process by repeating the following steps until reaching the number of given training rounds or the model is converged: 

\begin{enumerate}
[noitemsep,topsep=0pt,parsep=0pt,partopsep=0pt,leftmargin=4\parindent]
\item [\textbf{Step 1}] \textbf{Client sampling.} The server samples from a set of participating clients $\mathcal{N}_t\subset\mathcal{N}$ satisfying pre-defined eligibility requirements. 

\item[\textbf{Step 2}] \textbf{Broadcast.} The server broadcasts the current global model $\boldsymbol{w}_{t-1}$ to the selected clients. $t$ is the communication round index, where $t=1,\cdots, T$. 


\item[\textbf{Step 3}] \textbf{Local training.} 
Each selected client $j\in\mathcal{N}_t$ performs local training using its own dataset $\mathcal{D}_j$:
$$\label{eq:local-update}
\boldsymbol{w}_{t, j}(k)\leftarrow \boldsymbol{w}_{t, j}(k-1)-\eta \nabla \mathcal{L}_j(\boldsymbol{w}_{t, j}(k-1),\mathcal{D}_j),
$$
where $\eta$ is the learning rate and $k=1,\cdots,K$ is the index of local iterations.  

\item[\textbf{Step 4}] \textbf{Global aggregation.} The FL server obtains a new global model $\boldsymbol{w}_t$ by weighted-averaging the local models collected from the selected clients in round $t$:
$$
\boldsymbol{w}_t \leftarrow \sum_{j\in\mathcal{N}_t} \frac{|\mathcal{D}_j|}{|\cup_{j\in\mathcal{N}_t}\mathcal{D}_j|}\boldsymbol{w}_{t, j}(K). 
$$
\end{enumerate}

Unlike traditional centralized learning, where all training data is centralized on a server or a cluster, 
FL presents the following unique challenges.
(1) \textbf{Data heterogeneity}: 
in FL, local data across clients is often heterogeneous, varying in both labels and volumes. 
(2) \textbf{Loose connectivity}: FL clients may frequently join or leave the network, experiencing varying levels of latency and bandwidth due to fluctuating availability and network conditions. 
(3) \textbf{System heterogeneity}: 
FL involves diverse hardware and software environments—ranging from IoT devices to smartphones and vehicles, and from Android to iOS. 
These fundamental differences between centralized learning and FL drive the need to systematically explore the unique characteristics of FL training for ML. 
\section{Methodology}
\label{sec:method}

This section describes our experimental methodology for a comprehensive investigation of FL impact on ML model accuracy, including the FL framework, critical FL knobs we consider, model and workload selection, as well as the hardware setup.

\heading{Framework.}
To obtain high-fidelity evaluation results, we chose Flower(v1.11)~\cite{flower, flower_web}, a popular FL framework as a unified platform to experiment various FL implementations and strategies.
It decouples the configurations of the server, clients, and communication.
The server employs the {\em Strategy} abstraction that performs global computations, e.g., parameter aggregation, and orchestrate the distributed learning process.
Local computations on actual data to train and evaluate ML model parameters are performed on clients with the {\em Virtual Client} abstraction, which can run on heterogeneous edge hardware platforms.
The server communicates with clients with the {\em Client Manager}, which can perform client sampling, and {\em Client Proxies}, each dedicated to a single client.
Messages between clients and the server are serialized via the {\em Flower Protocol}.

Each of the above components is highly customizable: {\em Strategy} supports arbitrary FL algorithms, {\em Virtual Clients} support arbitrary ML frameworks (e.g., Pytorch or TensorFlow Lite), and the messaging protocol supports arbitrary communication stacks and application semantics.
Hence, Flower provides the necessary flexibility for us to vary various FL configuration knobs as described in the next subsection.
Due to the virtual client abstraction, FL applications developed atop Flower can theoretically be deployed to real-world edge devices transparently to harness distributed data at large scale.


\heading{FL Configuration Knobs.}
Based on the overall architecture as shown in Figure~\ref{fig:fl_arch}, we investigate the impact of FL on model accuracy from different aspects, including data distribution, scale, sampling, local learning, and finally global federation.
These aspects differentiate FL from centralized learning, encompassing {\em data} (imbalanced data distribution across clients), {\em computation} (both local computation on individual clients and global computation on the server), and {\em distribution} (the number of clients involved in the learning process and how the server selects them).

\begin{table}[t]
\small
\setlength{\tabcolsep}{3pt}
    \caption{Aspects of FL and the configuration knobs. $\alpha$ and $\beta$ are the concentration parameters in Dirichlet distribution.}
    \centering\begin{tabular}{llc}
        \toprule
        {\bf FL Aspect} & {\bf Configuration Knob} & {\bf Range} \\
    	\midrule
    	\multirow{2}{*}{Data distribution} & Non-IID distribution & $\alpha \in$  [0.01, 100] \\
    	& Volume imbalance & $\beta \in$[0.01, 100]\\
    	\midrule
    	 Scale & Number of clients & [10, 1000] \\
        \midrule
    	 Sampling & Sampling rate & [25\%, 100\%] \\
        \midrule
    	 \multirow{2}{*}{Local learning} & Batch size & [4, 128] \\
    	& Epochs per round & [1, 50] \\
        \midrule
        Global federation & FL algorithm & FedAvg, FedAdam, FedYogi \\
        \bottomrule
    \end{tabular}
	\label{tab:fl_knobs}
\end{table}

Table~\ref{tab:fl_knobs} shows the configuration knobs that control FL behavior in each aspect, all of which can impact model accuracy. Details (implications on accuracy and how we vary them for each ML task) are described in the following sections.

\heading{Workloads, Models, and Metrics.}
Our workload selection covers four common data forms generated by various applications: texts, images, audios, and videos, for which a wide spectrum of ML models have been developed~\cite{transformer, cnn_origin, sora, chatgpt, wavenet, stab_diff}.
We select a popular ML task for each data type.
For texts, we perform next-token prediction using the Shakespeare dataset~\cite{shakespeare}, which contains all of Shakespeare's plays totaling over one million lines.
For audios, we conduct environmental sound classification using the ESC-50 dataset~\cite{esc50}, comprising two thousand labeled audio recordings for benchmarking sound classification.
For images, we perform image classification using the CIFAR-10 dataset~\cite{cifar10}, which includes 60,000 images across ten classes.
Finally, for videos, we perform human action recognition using the UCF101 dataset~\cite{ucf101}, featuring 13,000 videos across 101 action categories.
We reserve 20\% of each dataset for testing and distribute the rest to clients for FL.

We select {\em state-of-the-art} models to perform each of the above tasks.
For images, audios, and videos, we compare the performance of recent models from popular leaderboards~\cite{leaderboard_images, leaderboard_videos, leaderboard_audios} and adopt one of the most accurate models.
For the text task, given the popularity of large language models (LLMs) and their promising ability for text generation, we select a recent pre-trained LLM as our base model.
Table~\ref{tab:ml_workloads} shows the selected models.
Specifically, we use Alpaca 7B~\cite{alpaca}, which is a model fine-tuned from LLaMA 7B~\cite{llama}. 
LoRA~\cite{lora} is adopted to further fine-tune the model on each FL client using the local data---instead of fine-tuning the original model, each client trains and sends to the server a much smaller LoRA module (rank = 8).
For the audio task, we use pre-trained weights from a Data-efficient Image Transformer (DeIT) and pass the averaged classification tokens into an Audio Spectrogram Transformer (AST) for further training with client data ~\cite{ASTpaper}.
For images, we use PyramidNet~\cite{pyramidnet}, which applies a novel residual unit to a deep CNN model for accurate image classification.
Finally, for the video task we adopt transformer-based Video Masked Autoencoder (VideoMAE)~\cite{videomae}. 
We initialize the model with pre-trained parameters on the Kinetics-400 dataset using self-supervised learning, and then remove the decoder and add classification head to the encoder for further fine-tuning on the UCF101 dataset~\cite{ucf101}.
\begin{table}[t]
\small
\setlength{\tabcolsep}{5pt}
    \caption{Selected ML workloads.}
    \centering\begin{tabular}{c c c c c}
        \toprule
         & {\bf Text} & {\bf Audio} & {\bf Image} & {\bf Video}  \\
    	\midrule
    	 {\bf Model} & Alpaca 7B + LoRA & AST & PyramidNet & VideoMAE \\
        \midrule
    	 {\bf Dataset} & Shakespeare & ESC-50 & CIFAR-10 & UCF101 \\
        \bottomrule
    \end{tabular}
	\label{tab:ml_workloads}
\end{table}

The definition of accuracy is rather obvious for each of the tasks: given $T$ test samples, the model achieves an accuracy of $T'/T$, where $T'$ denotes the number of correct predictions.
We further evaluate model performance with short-term accuracy, which is the accuracy achieved after ten rounds of FL, and long-term accuracy, that achieved after fifty rounds---nearly every model converges after fifty rounds as we show shortly.
We term the former {\em fast-learnt accuracy} and the latter {\em best-learnt accuracy}.
To evaluate the accuracy overhead of FL, we compare its accuracy with the converged test accuracy of the corresponding model in centralized learning.


\heading{Hardware Setup.}
We focus on measuring model accuracy, and thus our results are independent of specific hardware platforms. 
Nevertheless, our experiments are conducted in the Narval and Cedar  clusters in Digital Research Alliance of Canada~\cite{compca}.
In Narval, we use nodes with dual AMD Milan 7413 CPUs at 2.65 GHz with 498 GB of RAM and a single NVIDIA A100 GPU with 40 GB of memory. 
In Cedar, we utilize the nodes equipped with dual Intel Silver 4216 Cascade Lake CPUs at 2.1 GHz, 187 GB of RAM, and a single NVIDIA V100 GPU with 32 GB of memory.

\section{Data Distribution}
\label{sec:distri}
\begin{figure*}
    \centering
    \begin{subfigure}{0.25\textwidth}
        \centering
        \raisebox{0.30cm}{\includegraphics[width=\textwidth]{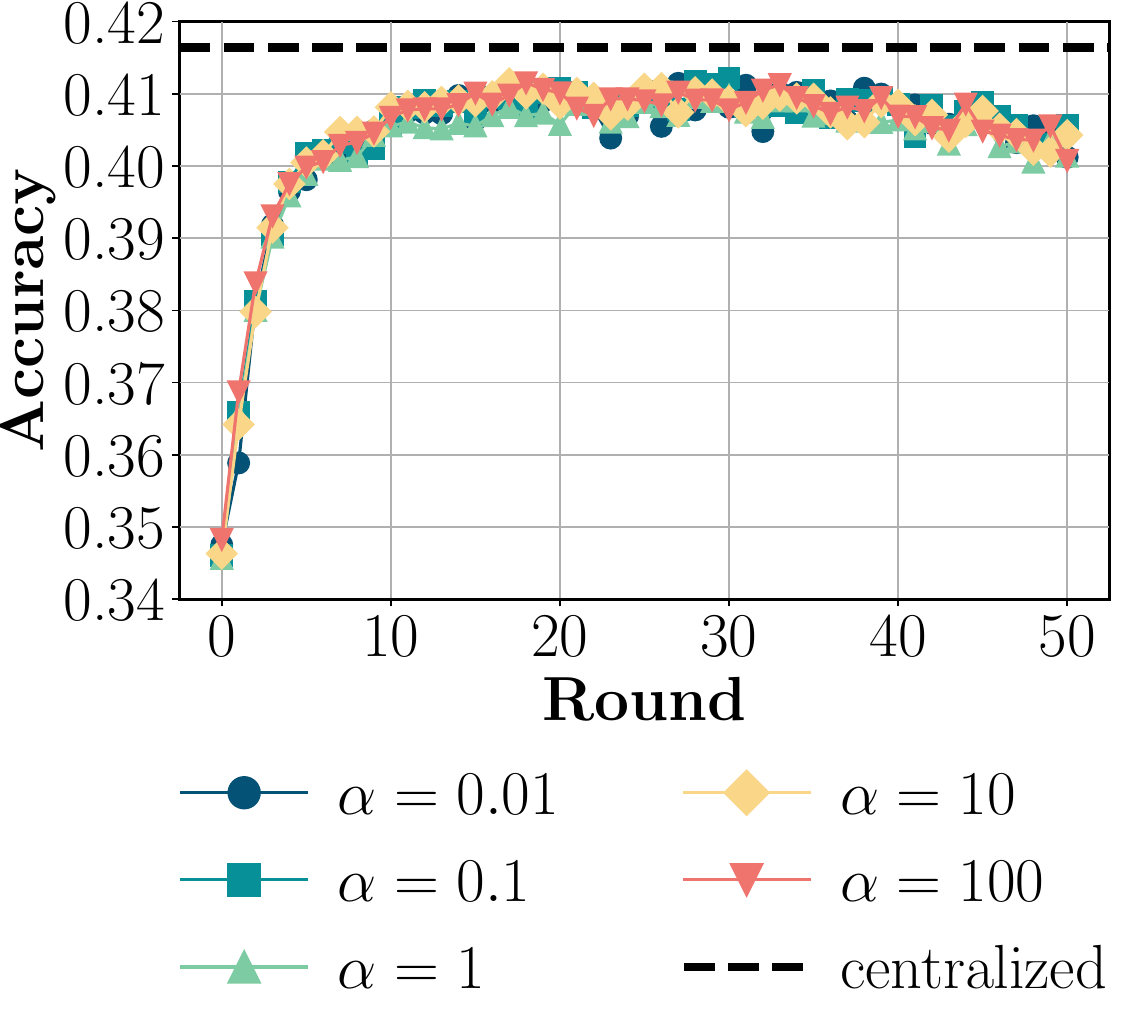}}
        \caption{Next-word prediction}
        \label{fig:non_iid_text}
    \end{subfigure}
    \begin{subfigure}{0.24\textwidth}
        \centering
        \raisebox{0.32cm}{\includegraphics[width=\linewidth]{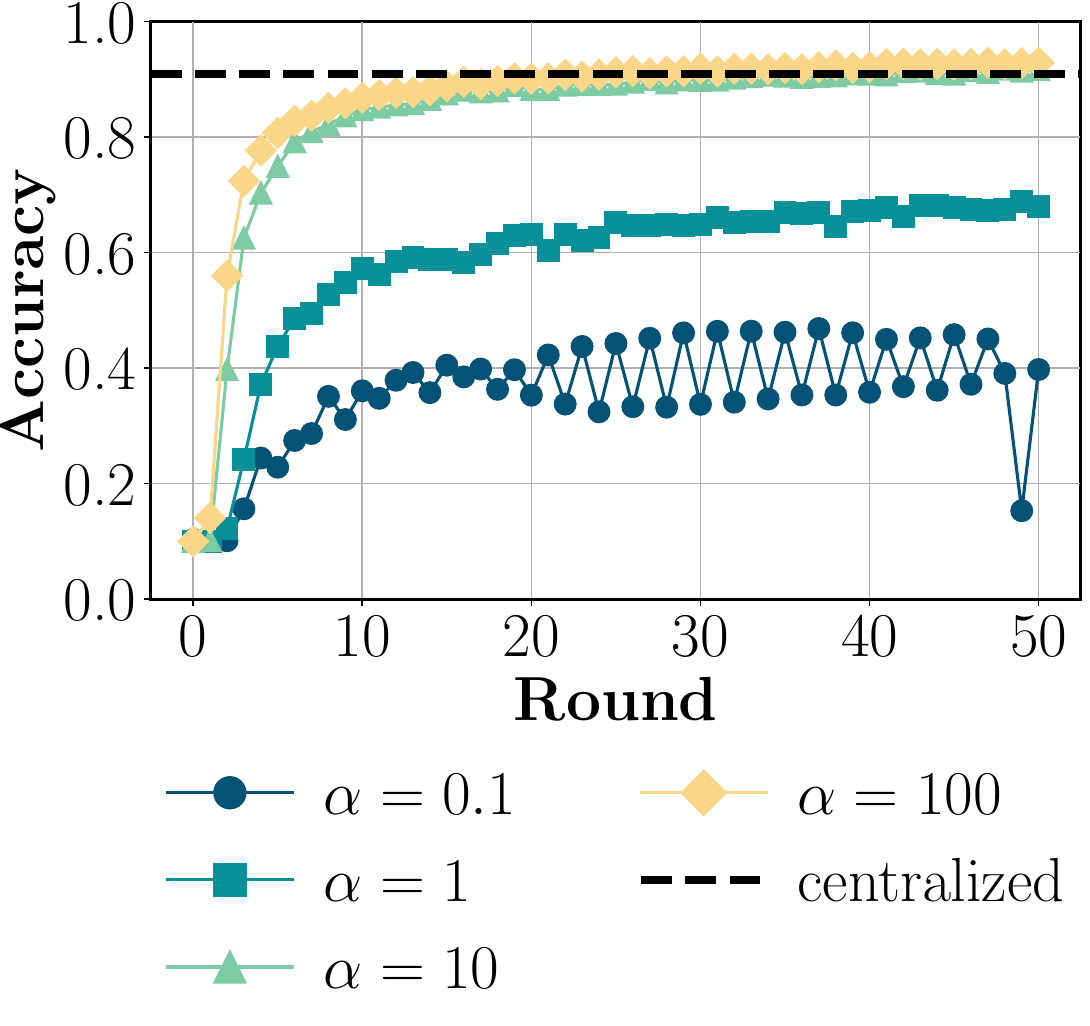}}
        \caption{Image classification}
        \label{fig:non_iid_image}
    \end{subfigure}
    \begin{subfigure}{0.25\textwidth}
        \centering
        \raisebox{0.00cm}{\includegraphics[width=\linewidth, height=4.35cm]{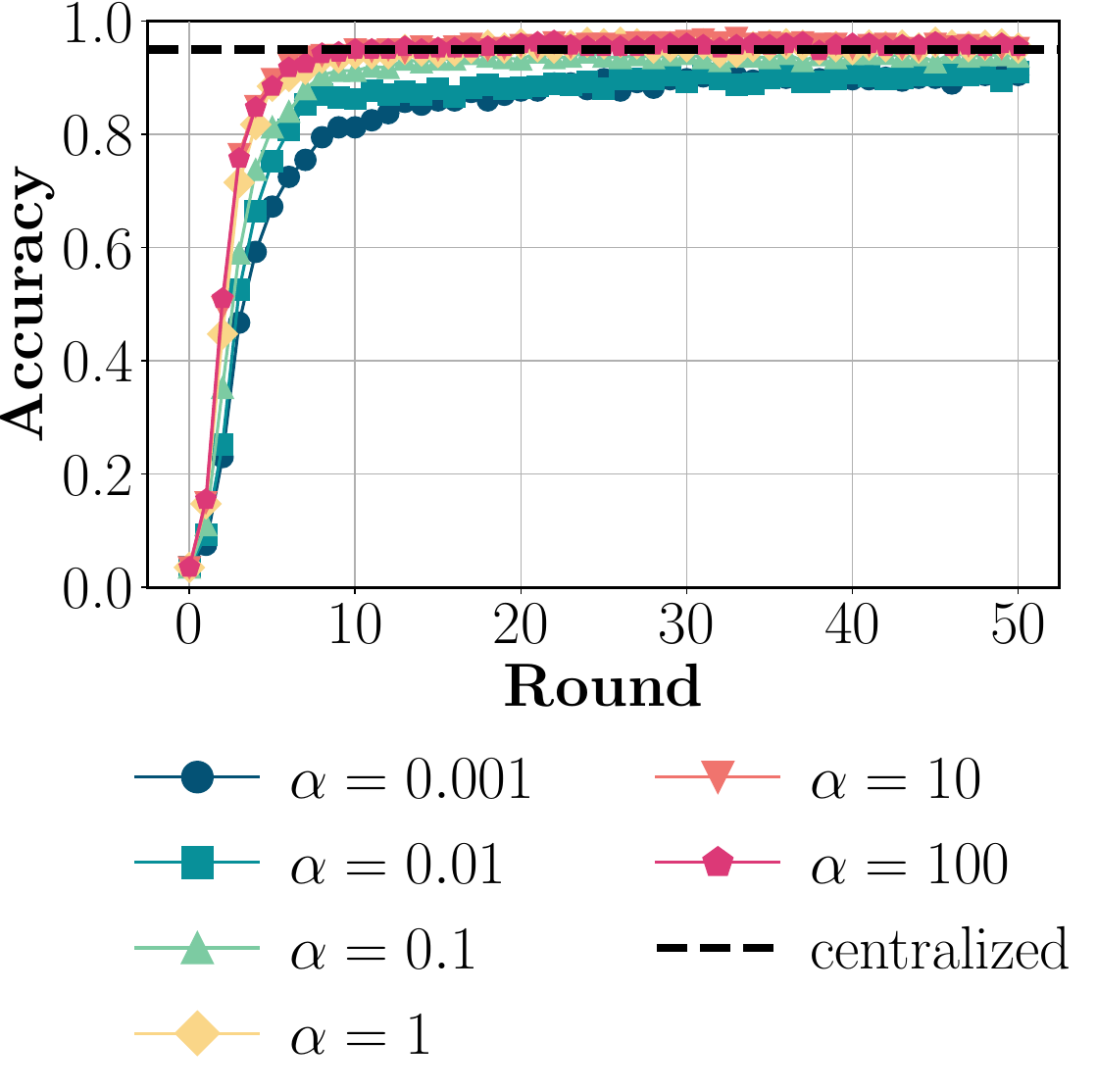}}
        \caption{Audio classification}
        \label{fig:non_iid_audio}
    \end{subfigure}
    \begin{subfigure}{0.24\textwidth}
        \centering
        \raisebox{0.36cm}{\includegraphics[width=\linewidth]{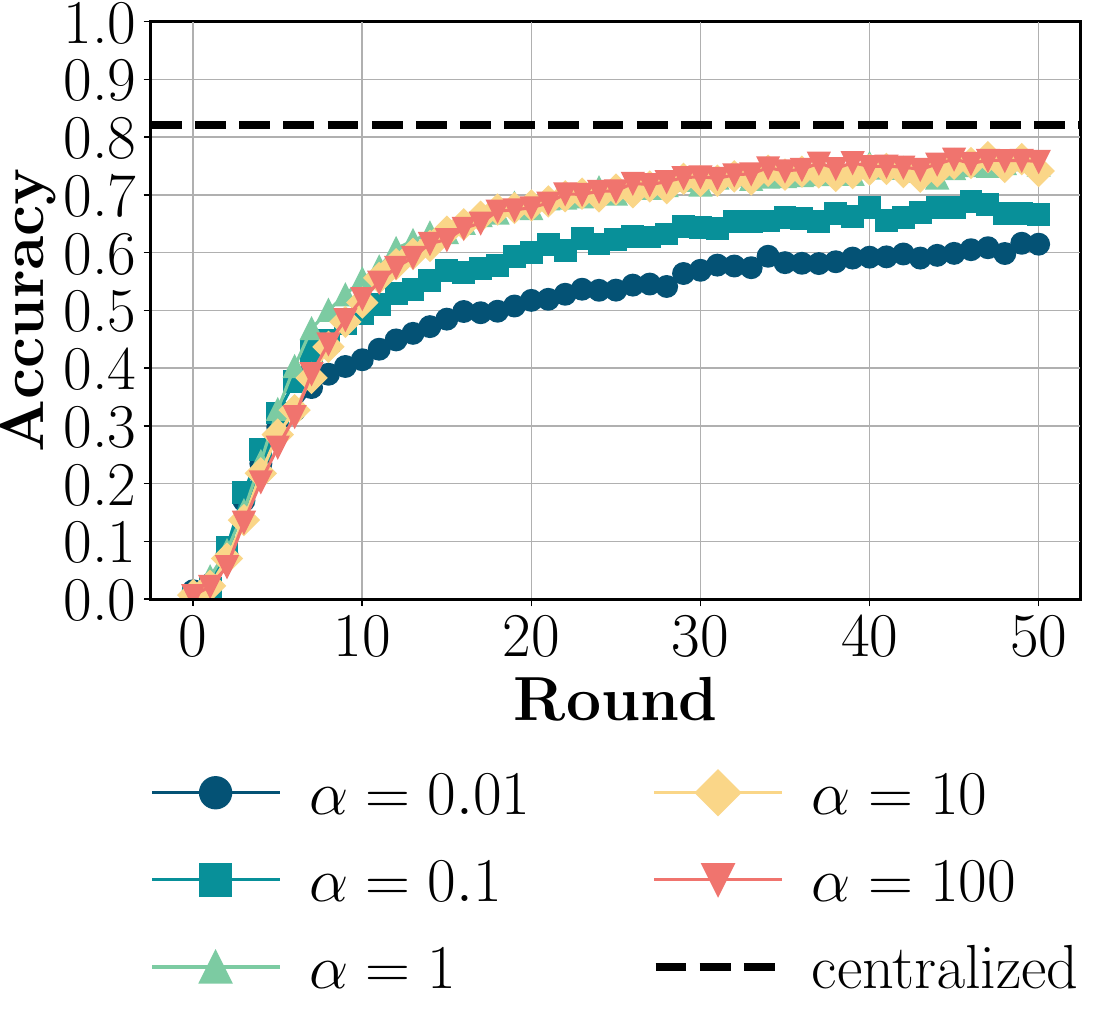}}
        \caption{Video action recognition}
        \label{fig:non_iid_video}
    \end{subfigure}
    \caption{Results for non-IID data distribution. The horizontal dashed line represents the accuracy of centralized learning.}
    \label{fig:non_iid}
\end{figure*}
\begin{figure*}
    \centering
    \begin{subfigure}{0.25\textwidth}
        \centering
        \raisebox{0.00cm}{\includegraphics[width=\linewidth]{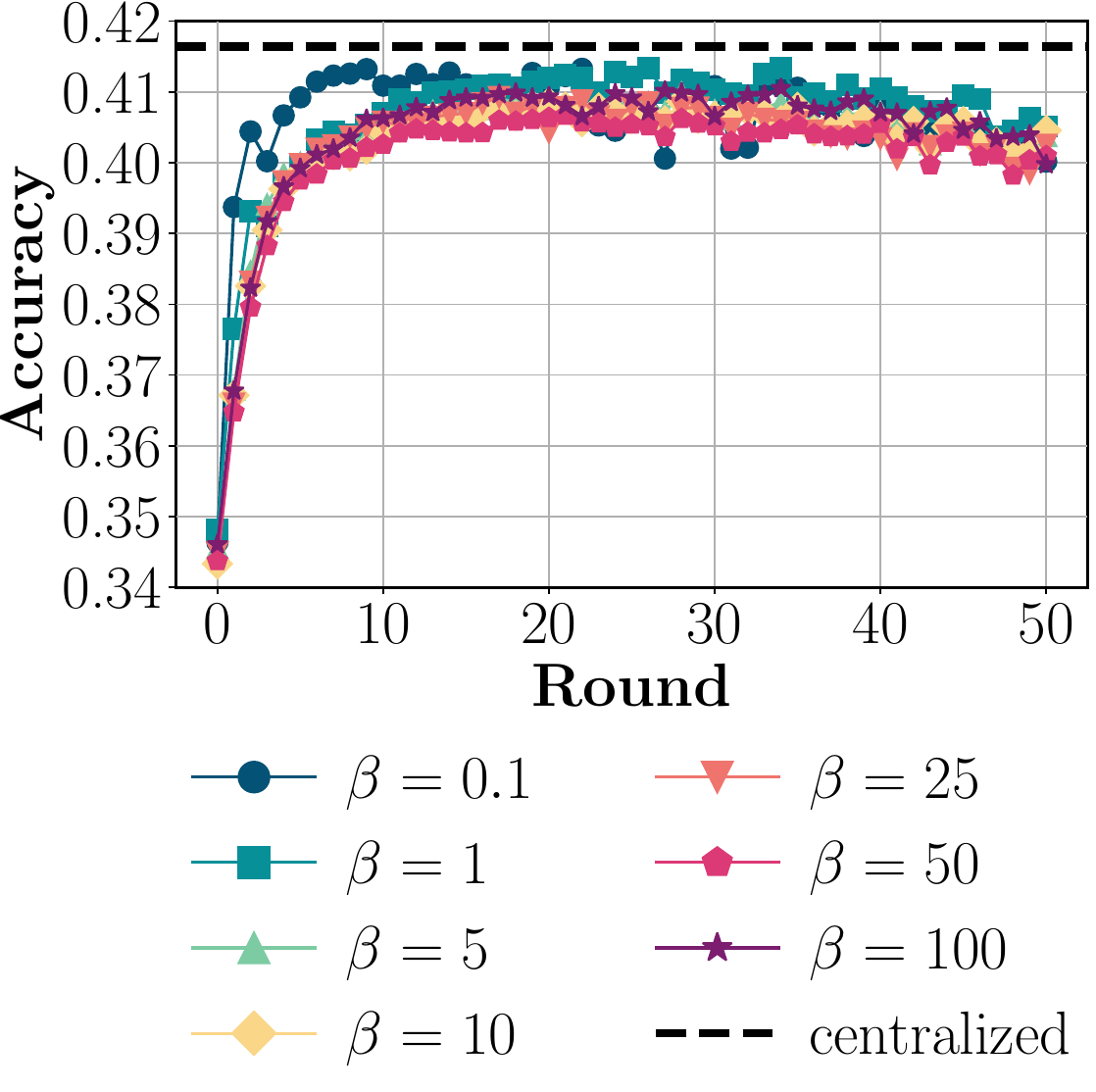}}
        \caption{Next-word prediction}
        \label{fig:quantity_text}
    \end{subfigure}
    \begin{subfigure}{0.24\textwidth}
        \centering
        \raisebox{0.02cm}{\includegraphics[width=\linewidth]{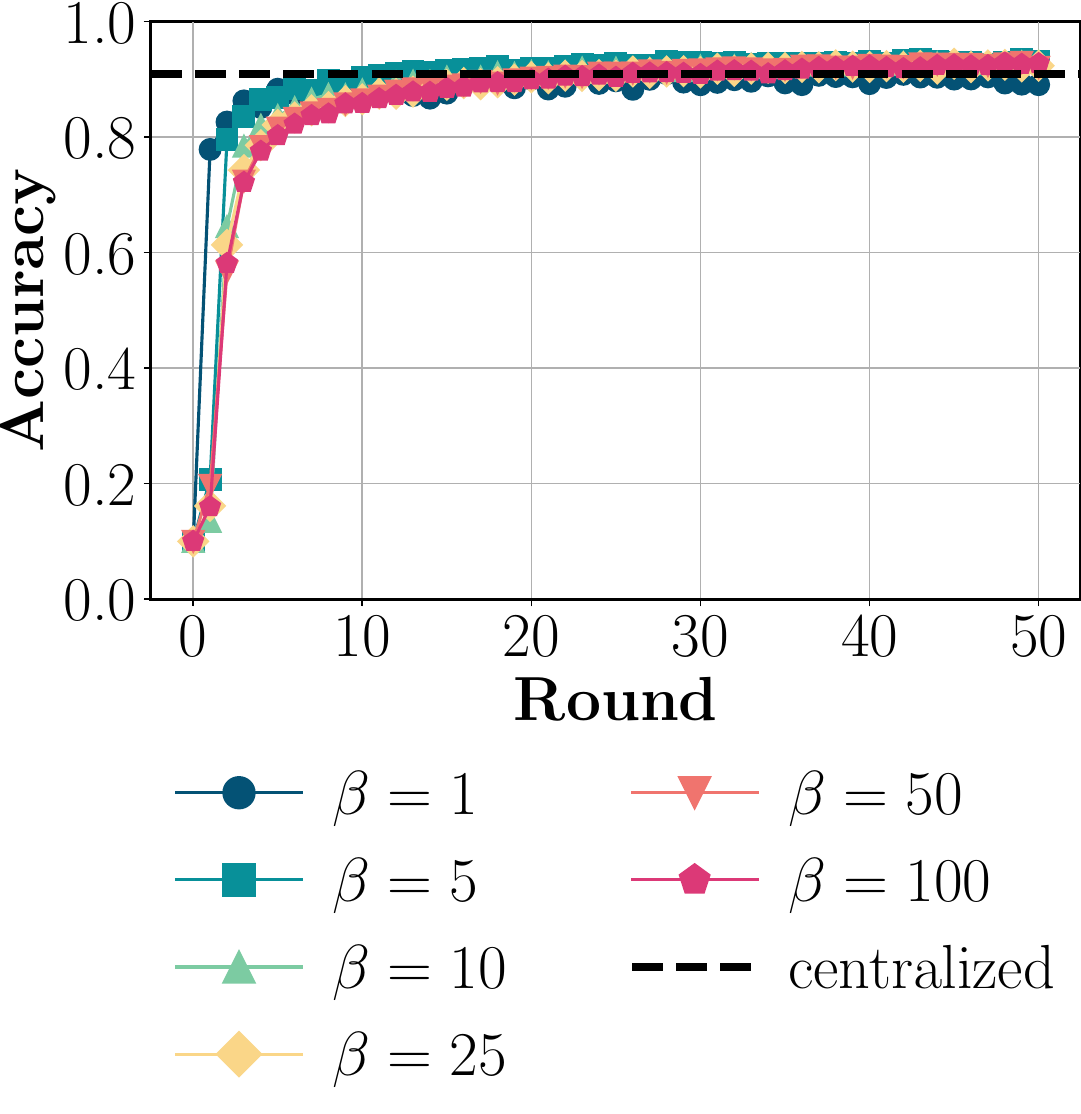}}
        \caption{Image classification}
        \label{fig:quantity_image}
    \end{subfigure}
    \begin{subfigure}{0.25\textwidth}
        \centering
        \raisebox{0.07cm}{\includegraphics[width=\linewidth, height=4.35cm]{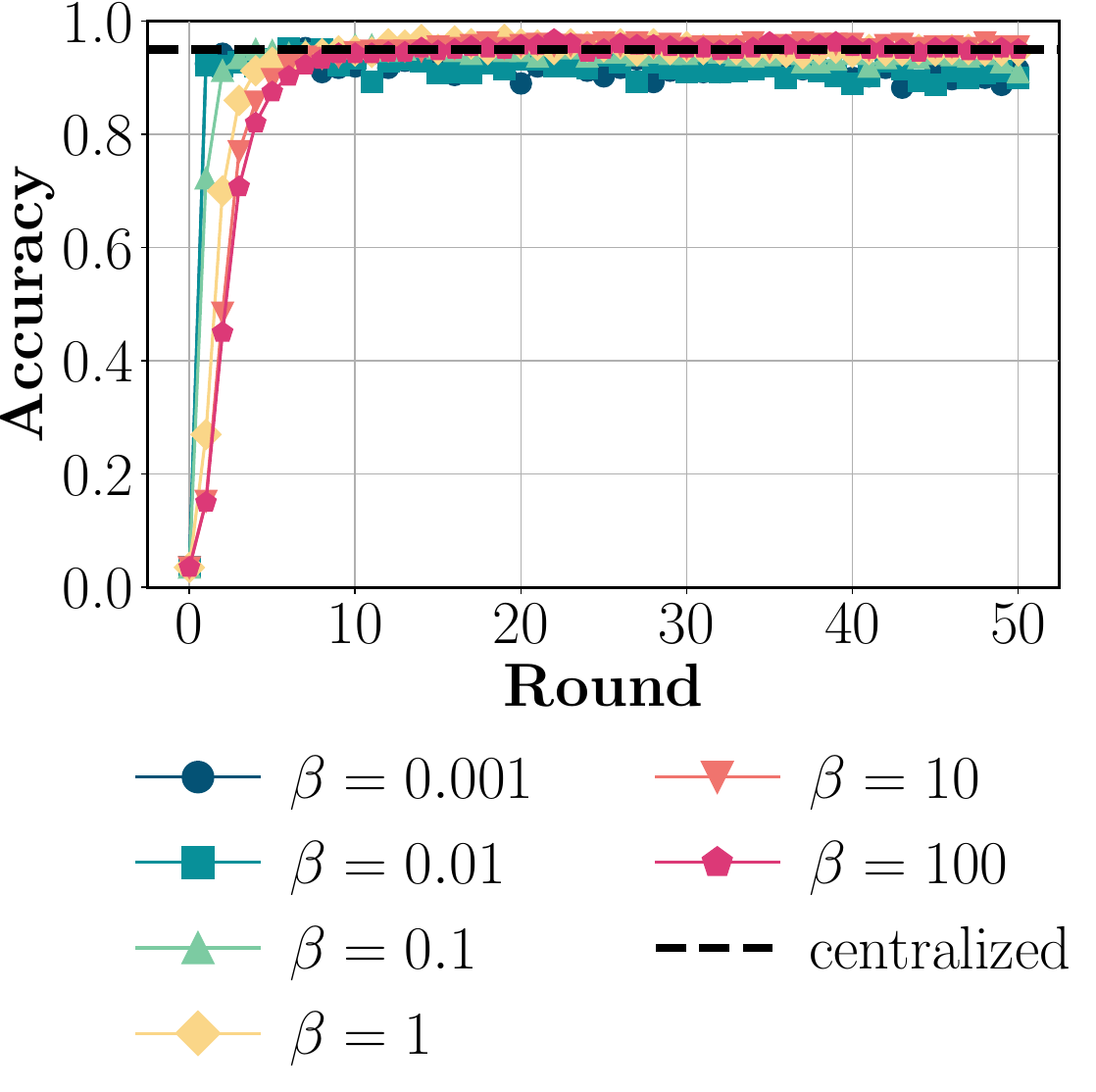}}
        \caption{Audio classification}
        \label{fig:quantity_audio}
    \end{subfigure}
    \begin{subfigure}{0.24\textwidth}
        \centering
        \raisebox{0.45cm}{\includegraphics[width=\linewidth]{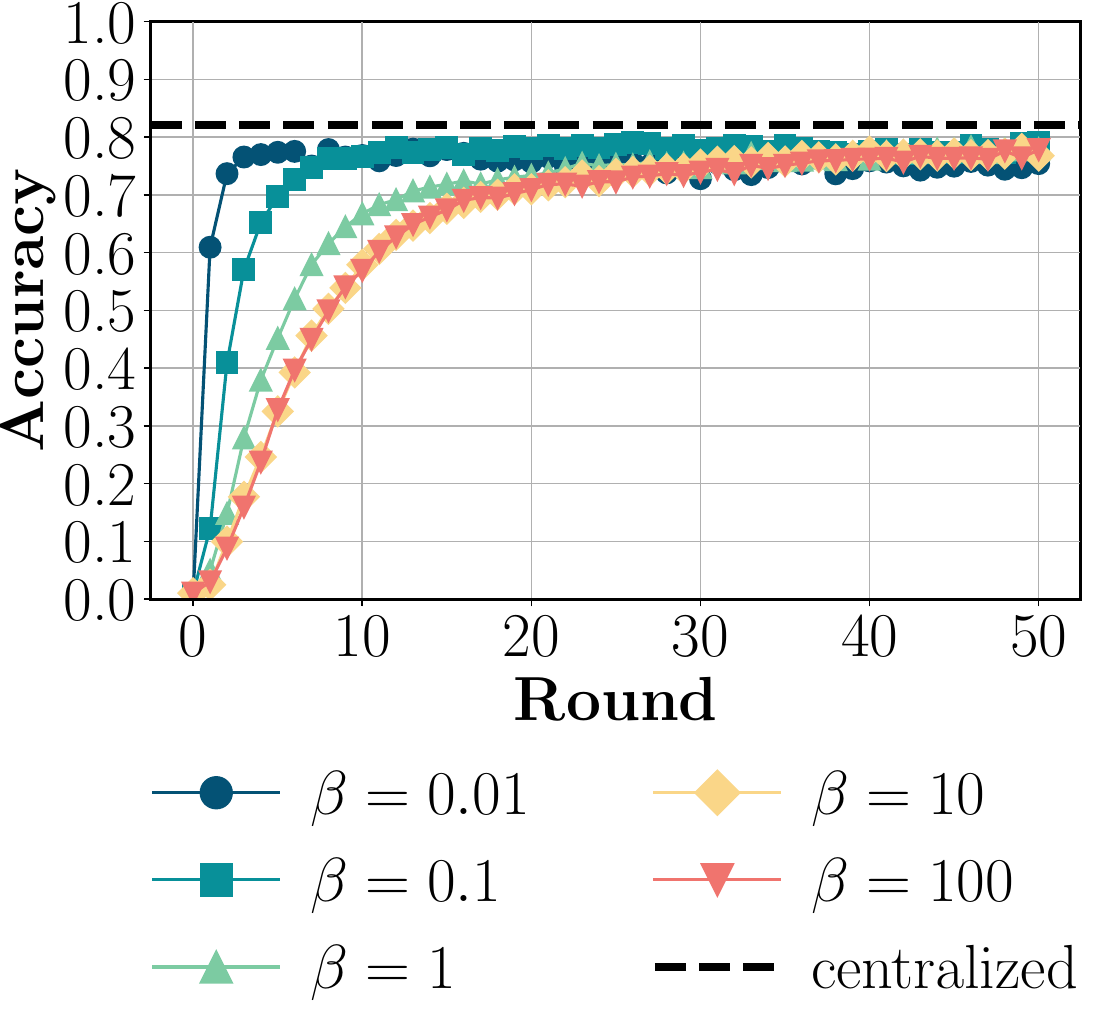}}
        \caption{Video action recognition}
        \label{fig:quantity_video}
    \end{subfigure}
    \caption{Results for skewed data volume distribution.}
    \label{fig:quantity}
\end{figure*}



Data distribution in federated settings is known to be unbalanced and skewed between clients~\cite{fl_origin}, i.e., classes of training data points are not independent and identically distributed, and some clients can generate significantly more training data than others.
We configure these two aspects of data distribution as follows.

\heading{Non-IID Data Distribution.}
For image, video, and audio data, 
we distribute label classes to clients via the Dirichlet distribution: $p_k \sim \text{Dir}_N(\alpha)$, where $N$ denotes the number of clients, $k$ the class, 
and $\alpha$ the concentration parameter that controls the degree of non-IID distribution (lower values mean higher degree of non-IID).
We assign $p_{k,n}$  of class $k$ to client $n$. 
Distributing the text data needs additional care, as samples and labels are less obvious.
Specifically, we segment each player's~\footnote{We use players to denote characters in the plays to differentiate characters in strings.} lines in Shakespeare's plays into multiple 80-token lines and label each line by the player.
We then distribute the lines using $p_f \sim \text{Dir}_N(\alpha)$, where $p_{f,n}$ represents the lines of player $f$ (in percentage) assigned to client $n$.
In addition, 
we also 
evaluate a natural splitting scheme, where all the lines of a player are assigned to a single client to emulate the scenario where different clients generate texts of distinct tones, styles, and contexts.

\heading{Volume Imbalance.}
To emulate the skewness of generated data volume between clients, 
we determine of the amount of training data distributed to clients with $q \sim \text{Dir}_N(\beta)$, where $\beta$ controls the degree of imbalance.
Client $n$ gets $q_n$ portion of the data. 

\subsection{Impact of Non-IID Distribution}


Figure~\ref{fig:non_iid} shows the detailed results, which we analyze below. 

\heading{Non-IID distribution can largely degrade model accuracy.}
We find that different degrees of non-IID distribution can lead to {\em drastically different} accuracy for certain tasks.
Specifically, for image classification (Figure~\ref{fig:non_iid_image}), the best-learnt accuracy when $\alpha=0.1$ is 46.9\%, which is much lower than that when $\alpha=100$ (92.9\%).
This finding also applied to the fast-learnt accuracy: 35.1\% when $\alpha=0.1$ vs. 85.9\% when $\alpha=100$.
This implies non-IID data distribution can be a major roadblock to model performance when deploying FL.

\heading{Different tasks have different sensitivity to the impact.}
The impact of this configuration differs greatly across tasks.
For instance, while the impact is pronounced for image classification, next-word prediction is nearly unaffected (Figure~\ref{fig:non_iid_text}).
It also varies between tasks: for audio classification, while the fast-learnt accuracy decreases from 94.5\% to 81.25\%, the best-learnt accuracy only drops from 96.5\% to 91\%, when the degree of non-IID increases ($\alpha=100$ to 0.001).
The impact becomes more prominent for the video task, the best-learnt accuracy drops from 76.2\% to 61.6\%.

\heading{Models can tolerate some non-IID distribution without sacrificing accuracy.}
While trivial for the next-word prediction task, even for image classification, the accuracy remains close when varying $\alpha$ from 100 to 10 (although the impact becomes apparent when further decreasing $\alpha$). 
This can also be generalized to the other two ML tasks (for audio classification, accuracy remains even when $\alpha$ drops to 1).
This result shows that non-IID distribution only becomes a major player that affects ML accuracy in FL when the degree is high (generally, when $\alpha$ is below 10).

\heading{Compared to centralized learning.}
As shown in Figure~\ref{fig:non_iid}, when the data distribution across clients moves towards IID, the difference between centralized learning and FL can be minimized.
When $\alpha=100$, while there is still some gap between FL and centralized learning for text (41.2\% vs. 41.6\%) and video (76.2\% vs. 82.1\%) tasks, model accuracy is identical between these two learning paradigms for the classification of images and audios.

\subsection{Impact of Volume Imbalance}
Figure~\ref{fig:quantity} shows the detailed results, which we analyze below.

\heading{The impact of volume imbalance is limited.}
Compared to non-IID distribution, most tasks experience insignificant accuracy differences when varying $\beta$.
Specifically, for text, image, and audio predictions, the fast-learnt accuracy varies within 4\% (40.2\% - 41.3\%, 85.8\% - 89.7\%, and 93.8\% - 95.3\%, respectively), and the best-learnt accuracy varies within 2\% (40.7\% - 41.3\%, 91.0\% - 93.5\%, and 95.3\% - 96.8\%, respectively).
The greatest impact of volume skewness is observed in the video task, where fast-learned accuracy ranges from 53.9\% to 77.9\%.
The effect on best-learned accuracy in the video task is minimal, varying within 2\% (from 77.8\% to 79.1\%).

\heading{More skewed volume distribution facilitates model learning but leads to overfitting.}
When zooming in on the effect of volume skewness, given a task, it is generally true that a more skewed volume distribution can help obtain an accurate model in the short term.
For instance, in the next-word prediction task, when $\beta=0.1$ (higher skewness), the fast-learnt accuracy is 41.3\%, which is higher than 40.5\%, the fast-learnt accuracy of $\beta=1$.
This phenomenon can also apply to other tasks.
This is the opposite of the impact of non-IID distribution.
The underlying reason is that clients with larger data volumes contribute more to the weight aggregation, dominating the global model updates. As a result, high-volume skewness creates an effect similar to centralizing the learning process.

However, distributing data more evenly across clients can prevent overfitting.
In Figure~\ref{fig:quantity_text}, $\beta=1$ outperforms $\beta=0.1$ as the learning process advances (e.g., after 25 rounds).
For audio/video predictions, the best-learnt accuracy is 95.8\%/79.1\% vs.  96.8\%/78.3\% when $\beta=0.1$ vs. $\beta=1$.
This is also true for the image task, the accuracy of high $\beta$'s exceeds that of low $\beta$'s after 20 rounds.

\heading{Compared to centralized learning.}
The FL overhead for text, image, and audio tasks is low, which is consistent with the impact of non-IID distribution (when $\alpha$ is high).
The difference from non-IID data distribution is that even when varying $\beta$, the model accuracy is close to that of centralized learning.
There is a noticeable gap between FL and centralized learning for video action recognition (5\%), requiring FL optimizations to bridge the gap.

\section{Client Sampling}
\label{sec:sample}

\begin{figure*}
    \centering
    \begin{subfigure}{0.245\textwidth}
        \centering
        \raisebox{0.01cm}{\includegraphics[width=\linewidth]{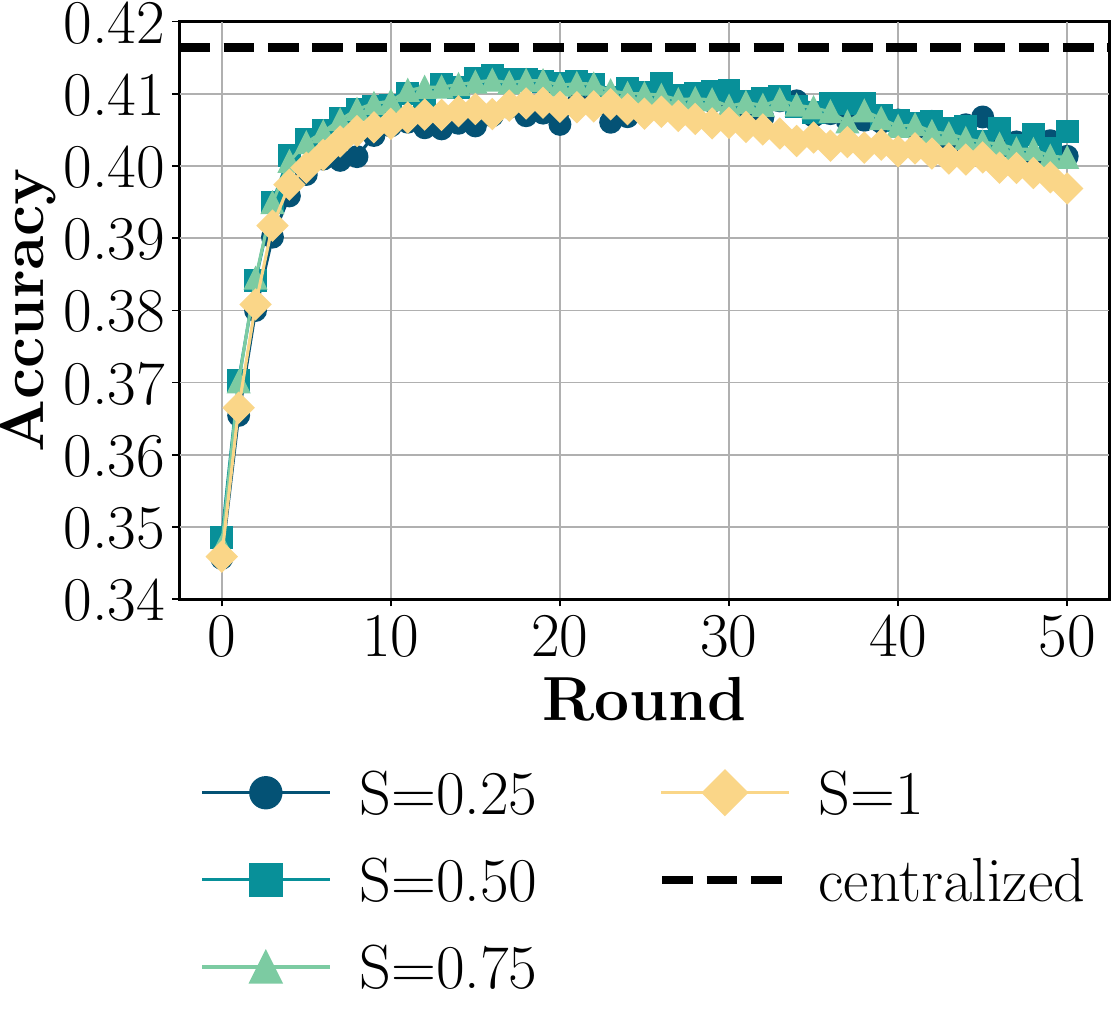}}
        \caption{Next-word prediction}
        \label{fig:sample_text}
    \end{subfigure}
    \begin{subfigure}{0.24\textwidth}
        \centering
        \raisebox{0cm}{\includegraphics[width=\linewidth]{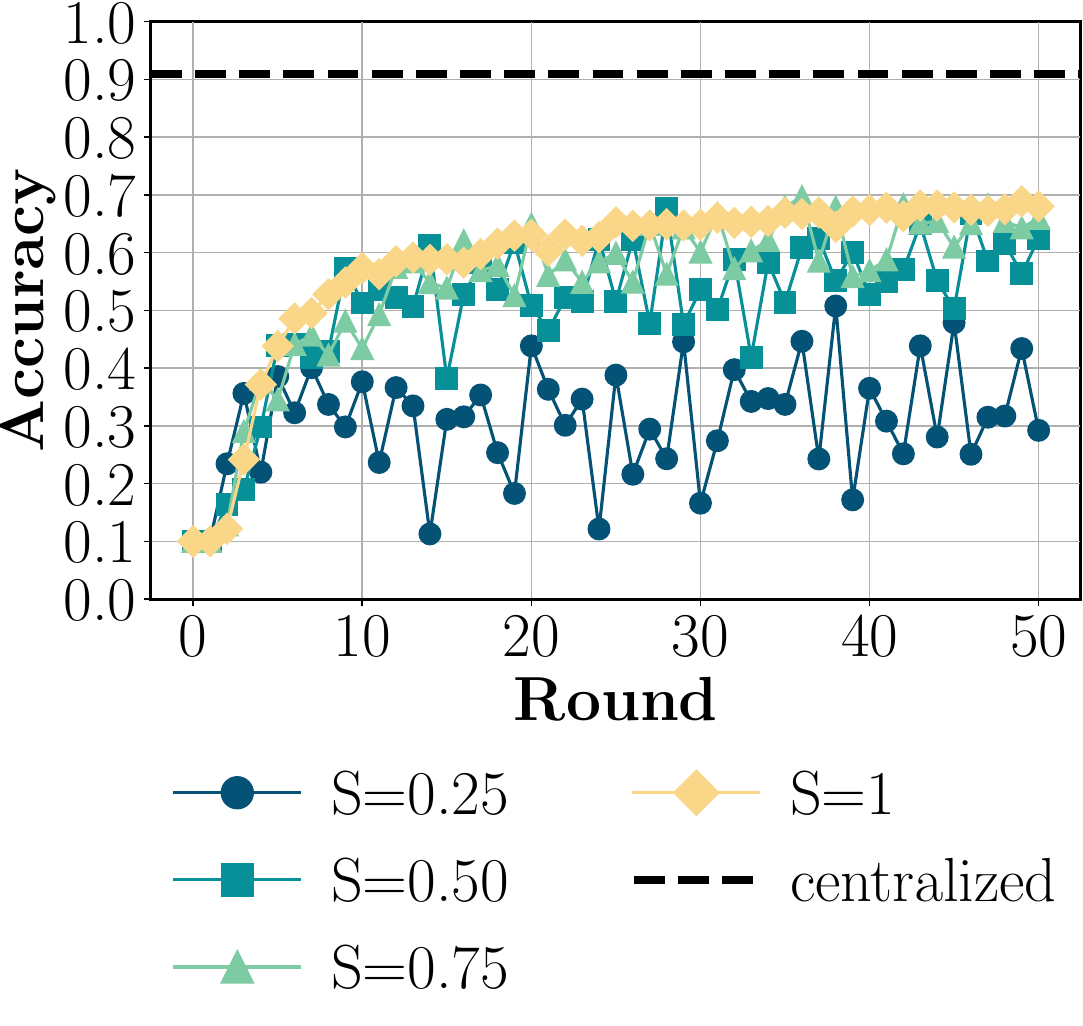}}
        \caption{Image classification}
        \label{fig:sample_image}
    \end{subfigure}
    \begin{subfigure}{0.24\textwidth}
        \centering
        \raisebox{0cm}{\includegraphics[width=\linewidth]{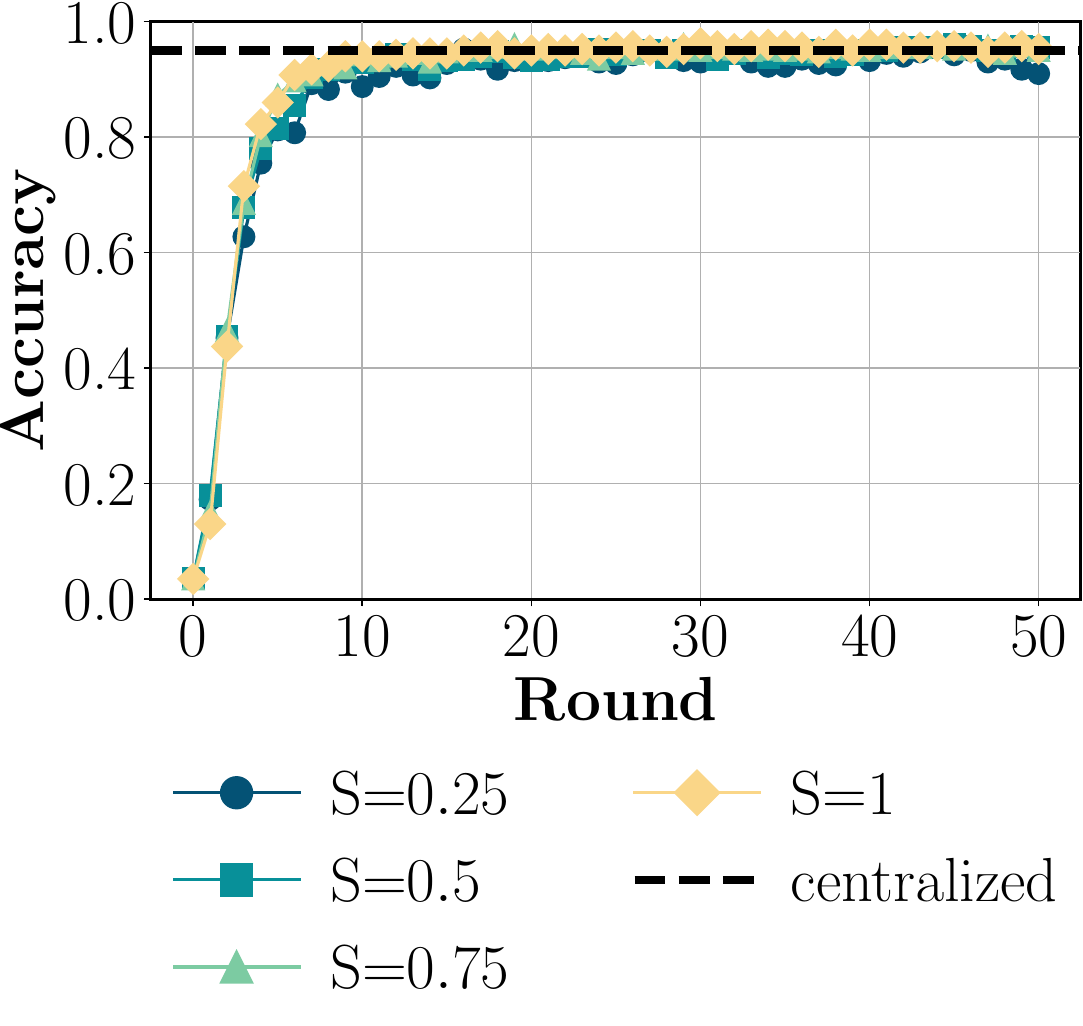}}
        \caption{Audio classification}
        \label{fig:sample_audio}
    \end{subfigure}
    \begin{subfigure}{0.24\textwidth}
        \centering
        \raisebox{0cm}{\includegraphics[width=\linewidth]{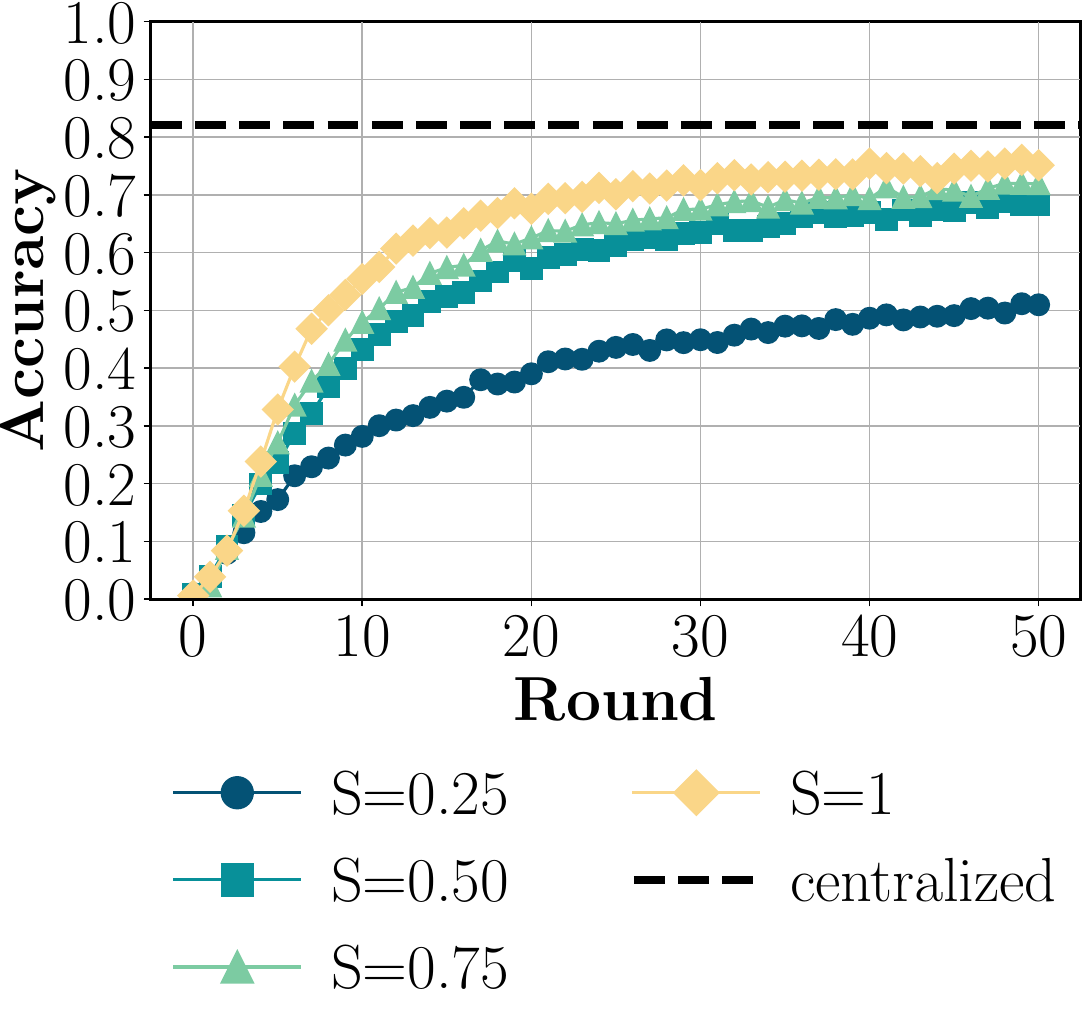}}
        \caption{Video action recognition}
        \label{fig:sample_video}
    \end{subfigure}
    \caption{Results for varying client sampling rate. `S' denotes the sampling rate.}
    \label{fig:sample}
\end{figure*}

As described in Section~\ref{sec:background}, the first step in an FL training round is client sampling---some clients may be excluded from the coming round of training. 
This is often applied to minimize the communication and computational cost, in the hope that the selected clients can still represent the global data distribution~\cite{oort, fedsamp}.
To understand the relationship between client sampling rate and model accuracy, 
we first fix the total number of clients and distribute the data during initialization, 
then vary the sampling rate (the fraction of clients that are randomly sampled) from 25\% to 100\% (no sampling) across experiments. 
As the sampling rate increases, more clients will be selected and contribute more training data to the training. 

\subsection{Impact of Client Sampling}

Figure~\ref{fig:sample} shows the detailed results, which we analyze below. 

\heading{When a significant fraction of clients is selected, the impact of client sampling is minimal.}
When more than 50\% of clients are sampled, the accuracy degradation is minimal across all tasks.
Varying the sampling rate barely affects accuracy for next-word prediction and audio classification as shown in Figures~\ref{fig:sample_text} and~\ref{fig:sample_audio}: 40.9\% and 96.3\% respectively for the text task and the audio task when all clients are selected vs. 41.3\% and 95\% when 50\% of clients are selected.
The impact of sampling is slightly higher on image and video classifications, but still in image classification (Figure~\ref{fig:sample_image}), the same metric only decreases from 68.9\% to 67.6\%; and in video action recognition (Figure~\ref{fig:sample_video}), the best-learnt accuracy changes from 76.1\% to 68.9\%.
This implies that client sampling can be an effective approach to saving resources and improving training speed for FL without significant accuracy reductions (if a sizeable portion of clients, e.g., 50\% of them, are selected).

\heading{Sampling less than half of clients can be a major factor of FL overhead.}
Our experiments further reveal that selectively sampling clients that participate in FL can cause noticeable model accuracy degradation when the sampling rate is below 50\%.
For instance, in image classification (Figure~\ref{fig:sample_image}), the best-learnt accuracy decreased from 68.9\% when all clients were involved to 50.8\% with a sampling rate of 25\%.
Similarly, in video action recognition (Figure~\ref{fig:sample_video}), the best-learnt accuracy drops from 68.9\% to 50.8\% when sampling 25\% of clients in each training round.
In addition, having fewer sampled clients leads to unstable training with severe fluctuations in the test accuracy curve, which can be observed in Figure~\ref{fig:sample_image}. 

\heading{Different tasks have different sensitivity to the impact of sampling rates.}
When the sampling rate is lower than 50\%, the impact of client sampling varies across tasks.
As we discussed, image and video predictions experience high accuracy degradation, which does not generalize to next-word prediction and audio classification.
This indicates that the optimal sampling rate, which balances sampling efficiency and model accuracy, is task-dependent. When applying aggressive client sampling to enhance learning speed, the rate should be tuned based on the ML task to reduce accuracy loss.

\heading{Compared to centralized learning.}
When the sampling rate is higher than 50\%, client sampling is not a contributor to FL accuracy overhead.
Although for the text, image, and video tasks, there is still a gap between centralized learning and FL (3.8\%, 22.1\%, and 6.1\% respectively with the best-learnt accuracy), the gap is caused by other FL aspects, e.g., the default non-IID data distribution.
As the impact of client sampling diversifies between tasks when the sampling rate is low, while image and video predictions with FL are much less accurate than centralized learning: 31\% and 40.2\% accuracy difference respectively with a sampling rate of 25\%, the text and audio tasks with FL achieve similar accuracy as centralized learning: only 0.8\% and 0.5\% accuracy difference, respectively. 
\section{Scale}
\label{sec:scale}
\begin{figure*}
    \centering
    \begin{subfigure}{0.25\textwidth}
        \centering
        \raisebox{0.29cm}{\includegraphics[width=\linewidth]{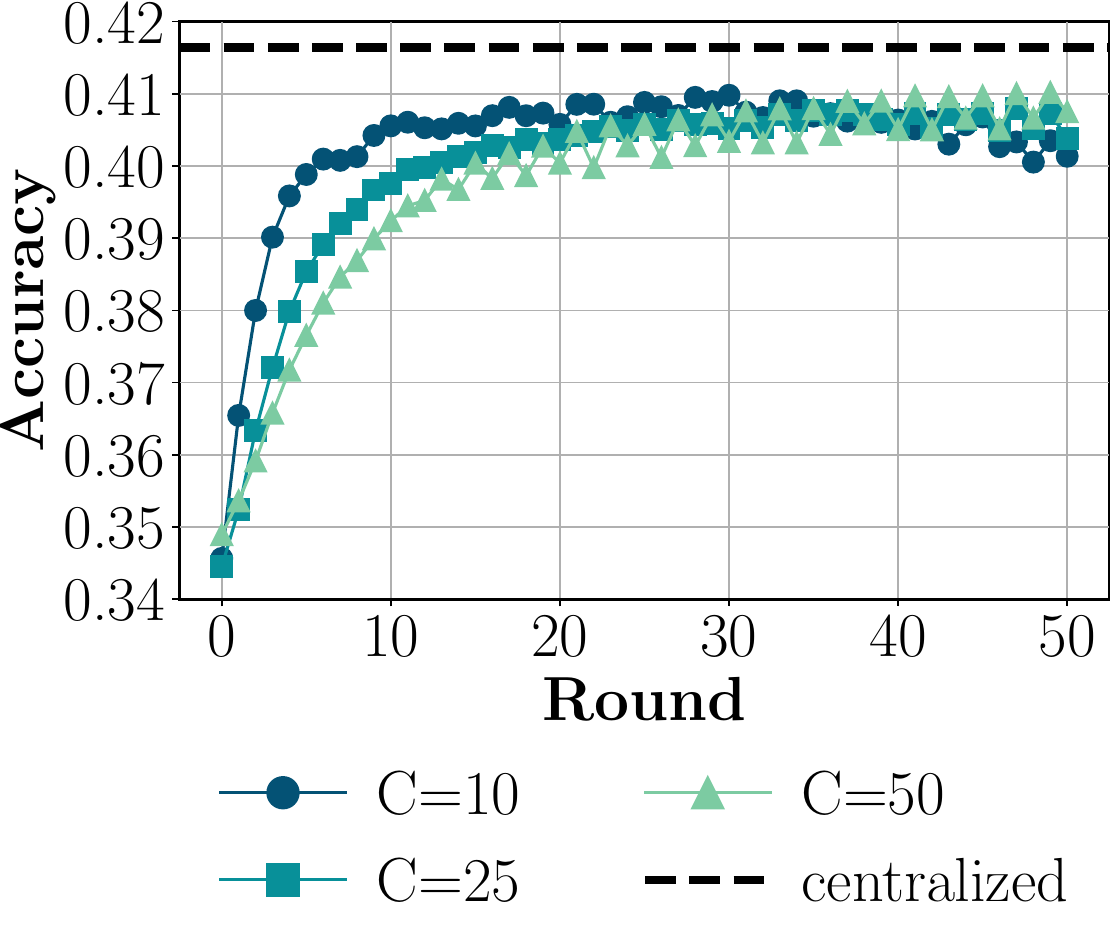}}
        \caption{Next-word prediction}
        \label{fig:scale_text}
    \end{subfigure}
    \begin{subfigure}{0.24\textwidth}
        \centering
        \raisebox{0.30cm}{\includegraphics[width=\linewidth, height=3.75cm]{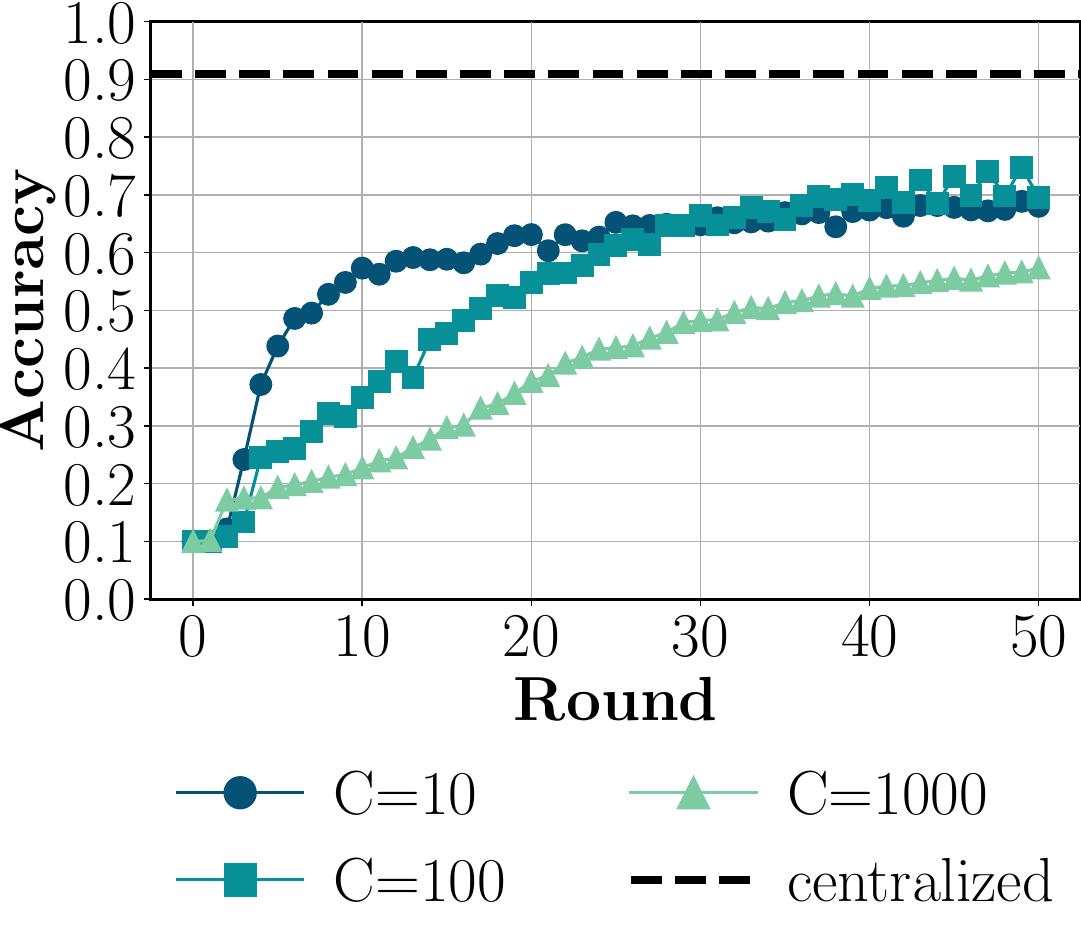}}
        \caption{Image classification}
        \label{fig:scale_image}
    \end{subfigure}
    \begin{subfigure}{0.24\textwidth}
        \centering
        \includegraphics[width=\linewidth]{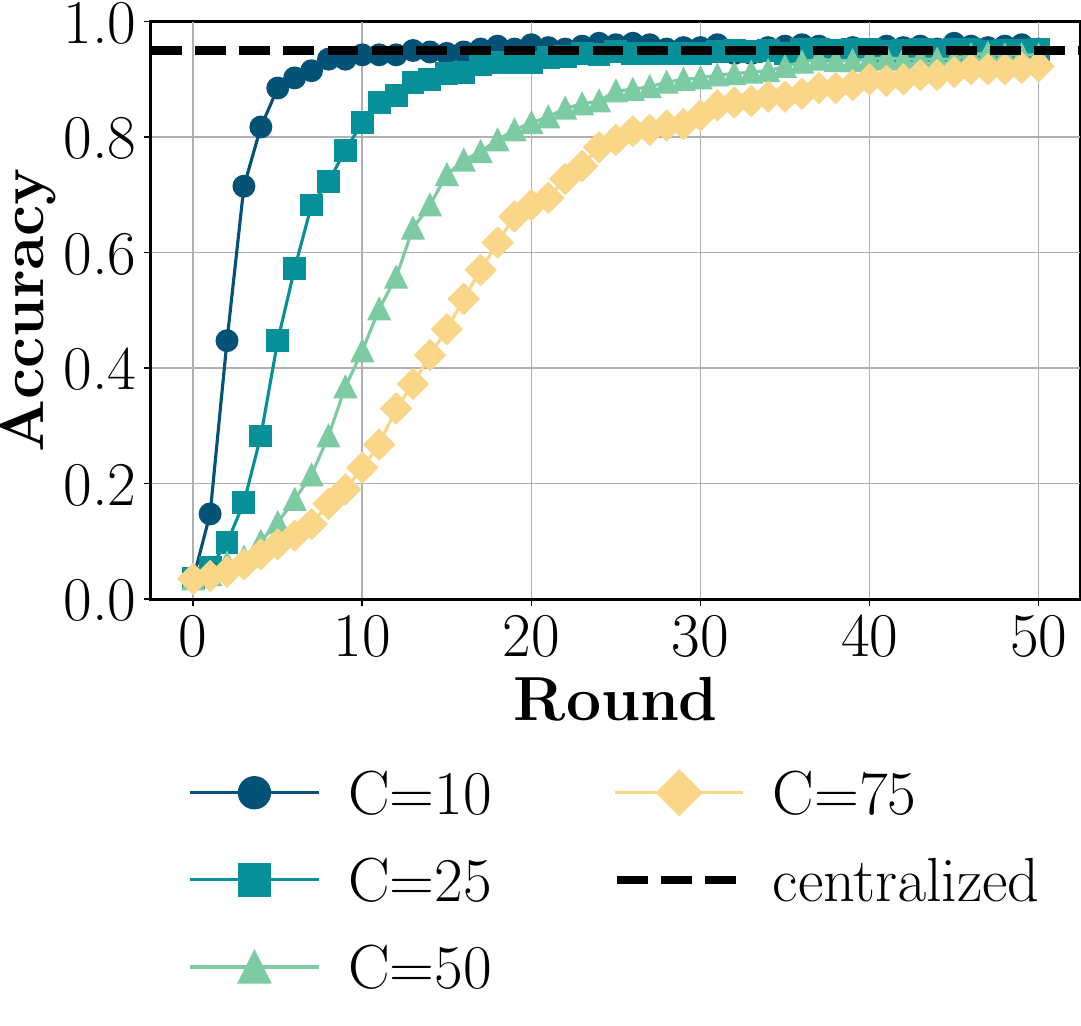}
        \caption{Audio classification}
        \label{fig:scale_audio}
    \end{subfigure}
    \begin{subfigure}{0.24\textwidth}
        \centering
        \raisebox{0cm}{\includegraphics[width=\linewidth]{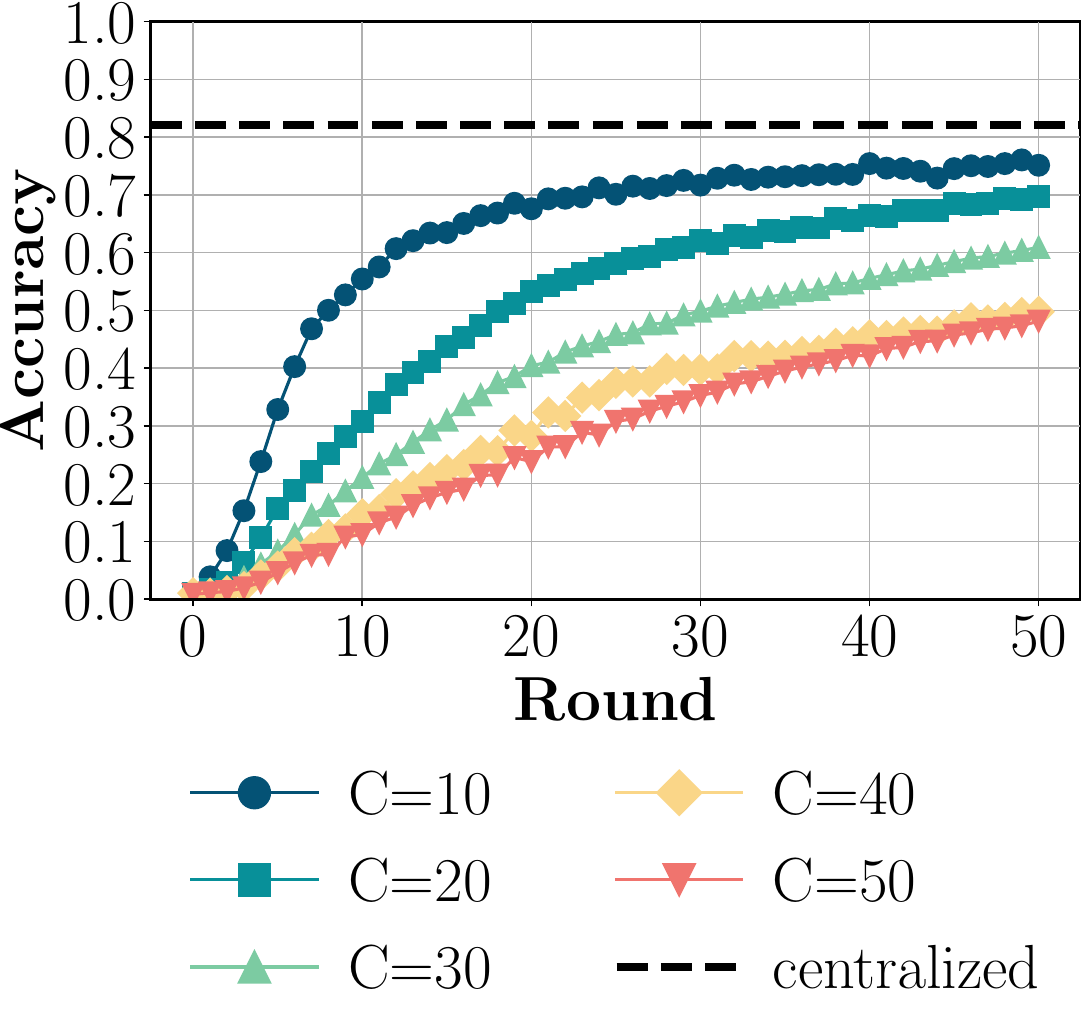}}
        \caption{Video action recognition}
        \label{fig:scale_video}
    \end{subfigure}
    \caption{Results for varying the scale of FL. `C' denotes the number of clients.}
    \label{fig:scale}
\end{figure*}


This configuration knob examines how the degree of decentralization impacts model accuracy, specifically the trade-off between increased participation and decreased individual contributions in FL. By maintaining the total dataset size and varying the number of clients, we emulate practical scenarios where FL is deployed across different numbers of edge devices, and observe changes in model accuracy as the scale of FL varies from 10 to 1000 clients.

\subsection{Impact of FL Scale}


Generally, the degree of distribution is among the most prominent factors manifested in FL overhead.
We detail our findings as follows.

\heading{Accuracy degradation with larger scales.}
Involving more clients in FL adds more challenges to short-term learning as 
information takes time to propagate.
Indeed, this phenomenon is observed across all tasks.
For image classification, where we have the smallest model, 
increasing the number of clients from 10 to 100 and eventually to 1000 degrades fast-learnt accuracy: 
49.5\%, 32.1\%, and 21.6\% (Figure~\ref{fig:scale_image}).
For next-word prediction and audio classification, the fast-learnt accuracy decreases from 10 to 50 clients (40.7\% vs. 39\%,  93.5\% vs. 36.8\%).
We also evaluate the effect of the number of clients on the video task.
As shown in Figure~\ref{fig:scale_video}, the fast-learnt accuracy decreases with increasing client size, achieving  52.7\%, 28.1\%, 18.7\%, 12.1\%, and 10.8\% for client sizes of 10, 20, 30, 40, and 50, respectively.

\heading{Diminished long-term impact.}
Perhaps surprisingly, 
the scale of FL has less impact as training rounds increase.
For instance, best-learnt accuracy for next-word prediction and audio classification is nearly the same for 10 and 50 clients (41.1\% vs. 41\%, 96.3\% vs. 94.5\%).
In image classification, the accuracy of 10 clients and 100 clients converges at 70\%.
While accuracy for 1000 clients remains lower than that of 100 and 10 clients after 50 rounds, it has not converged and still increases given more training rounds.
The observation can also be made in video action recognition (Figure~\ref{fig:scale_video}).
This result indicates that when deploying FL with many clients, allowing more training time is a safe strategy to acquire an accurate model.

\heading{Compared to centralized learning.}
The above findings can be compared with centralized learning.
FL overhead for short-term learning is high when there are a large number of clients: 2.7\%, 69.4\%, 76\%, and 29.4\% fast-learnt accuracy degradation with the largest number of clients, respectively, for the four tasks.
But when given sufficient training rounds, the overhead is minimized: only 0.6\%, 33.6\%, 2.8\%, and 6.1\% when observing best-learnt accuracy.
\section{Local Learning Strategies}
\label{sec:local}


We now evaluate two configuration knobs: batch size and epochs in this section.
Increasing batch size is an effective technique to leverage the high parallelism of GPUs, but large batch sizes can lead to poor generalization.
On one hand, small batch sizes essentially allow more frequent updates per epoch, resulting in faster convergence and better generalization. 
On the other hand, lower learning rates or careful regularization are required to improve learning stability for smaller batch sizes. 
To investigate how this knob may affect FL accuracy, we vary the batch size between 4 and 128 when performing local training on clients, while fixing other FL aspects. Another critical factor that affects local learning efficiency is how many epochs each client trains the model within each FL round.
More epochs allow clients to better exploit their local data, but may lead to overfitting issues.
To investigate the effect, we vary the number of epochs between 1 and 50 across different tasks.

\begin{figure*}
    \centering
    \begin{subfigure}{0.245\textwidth}
        \centering
        \includegraphics[width=\linewidth]{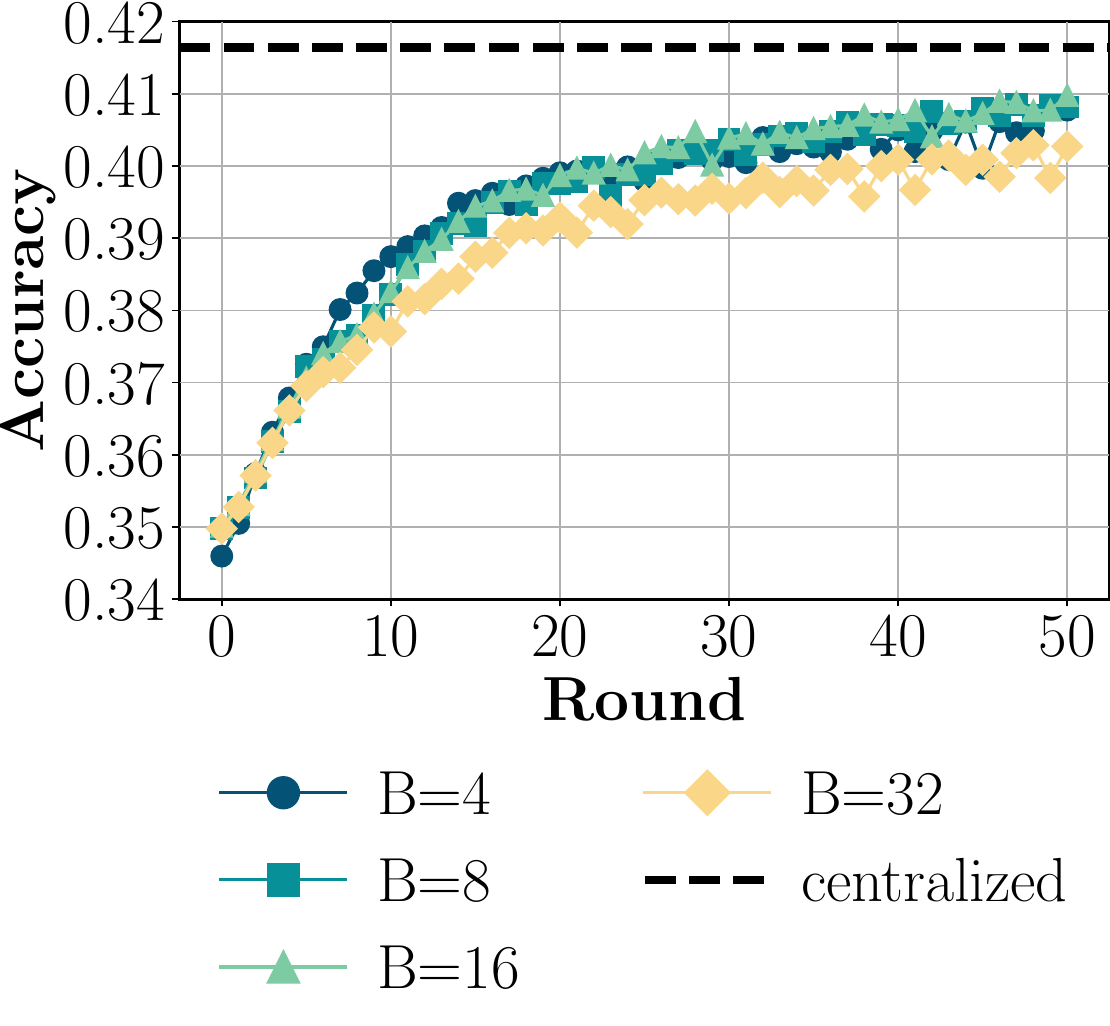}
        \caption{Next-word prediction}
        \label{fig:batch_text}
    \end{subfigure}
    \begin{subfigure}{0.24\textwidth}
        \centering
        \includegraphics[width=\linewidth]{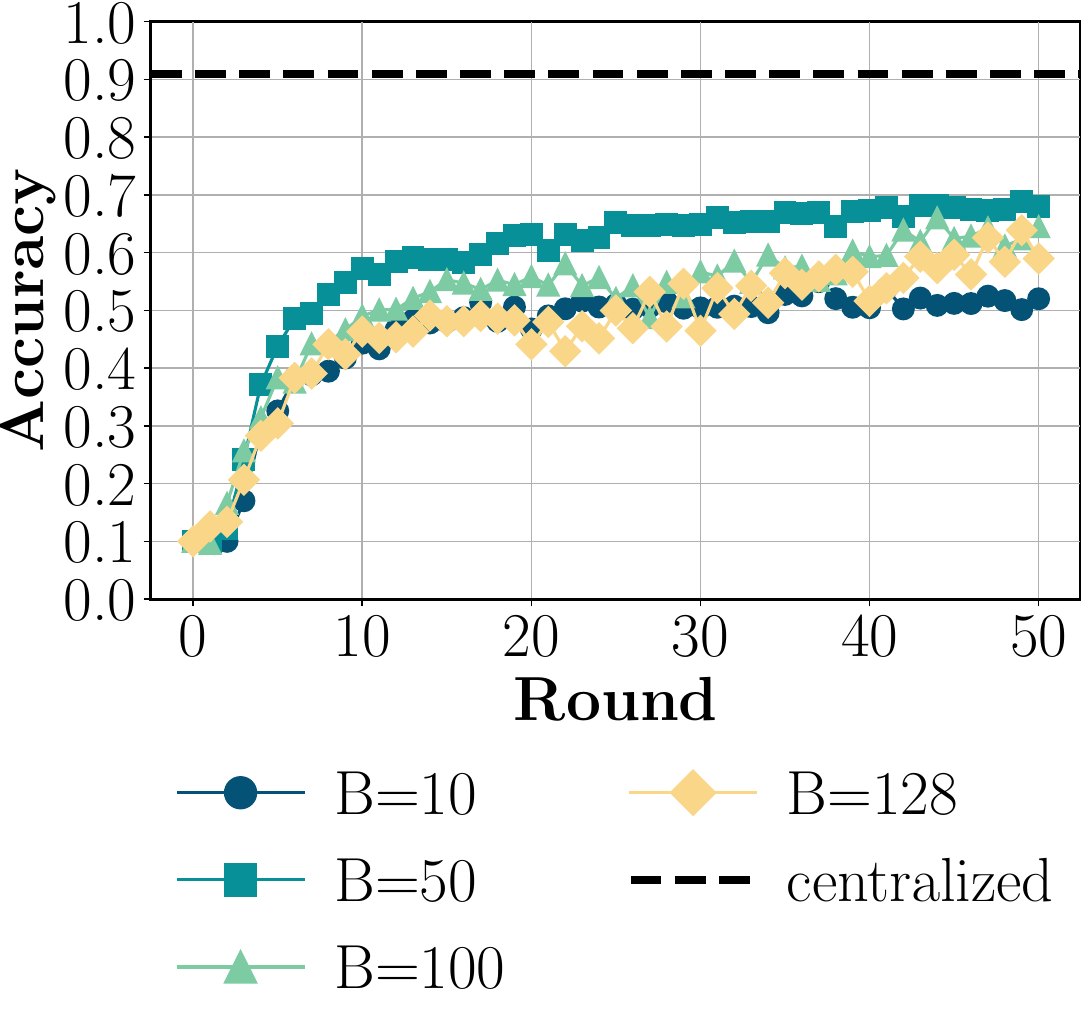}
        \caption{Image classification}
        \label{fig:batch_image}
    \end{subfigure}
    \begin{subfigure}{0.24\textwidth}
        \centering
        \includegraphics[width=\linewidth]{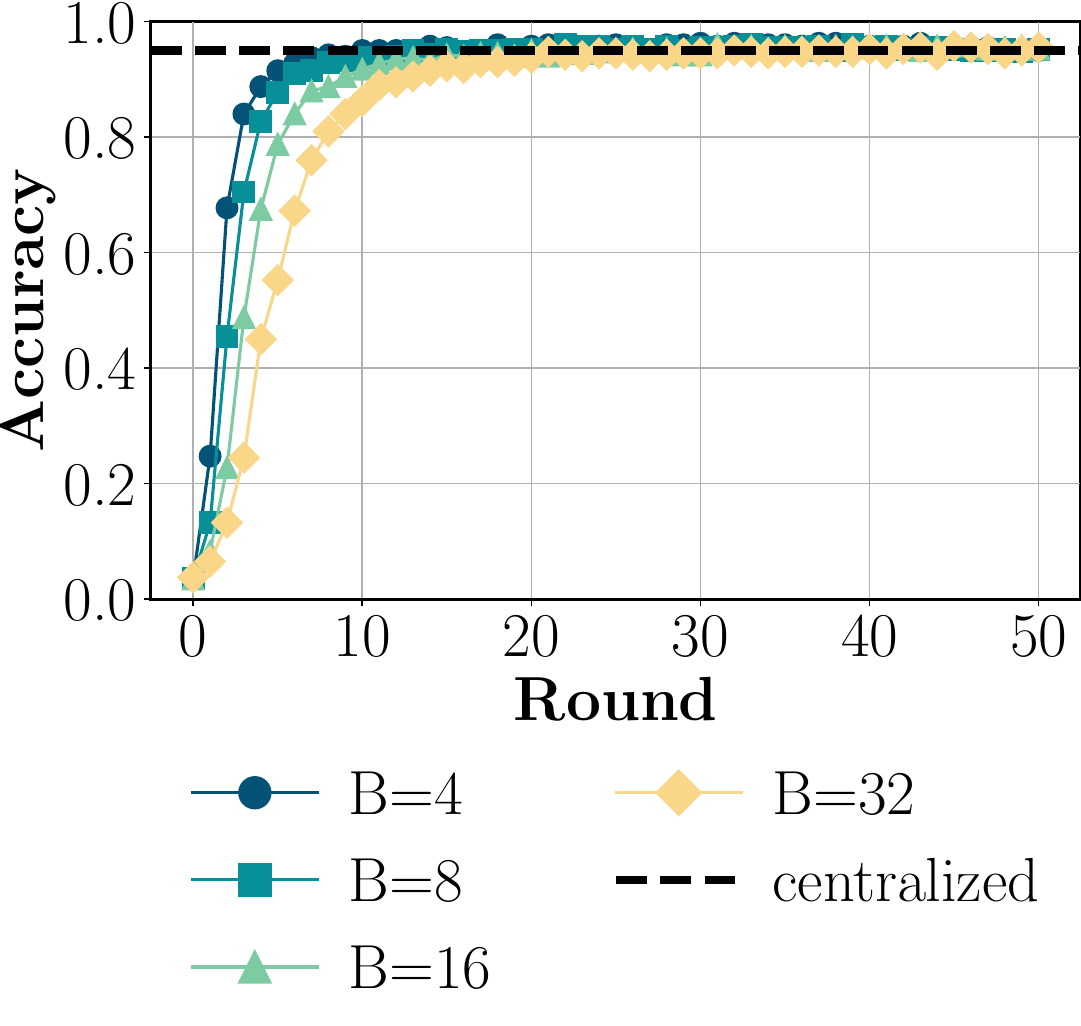}
        \caption{Audio classification}
        \label{fig:batch_audio}
    \end{subfigure}
    \begin{subfigure}{0.24\textwidth}
        \centering
        \includegraphics[width=\linewidth]{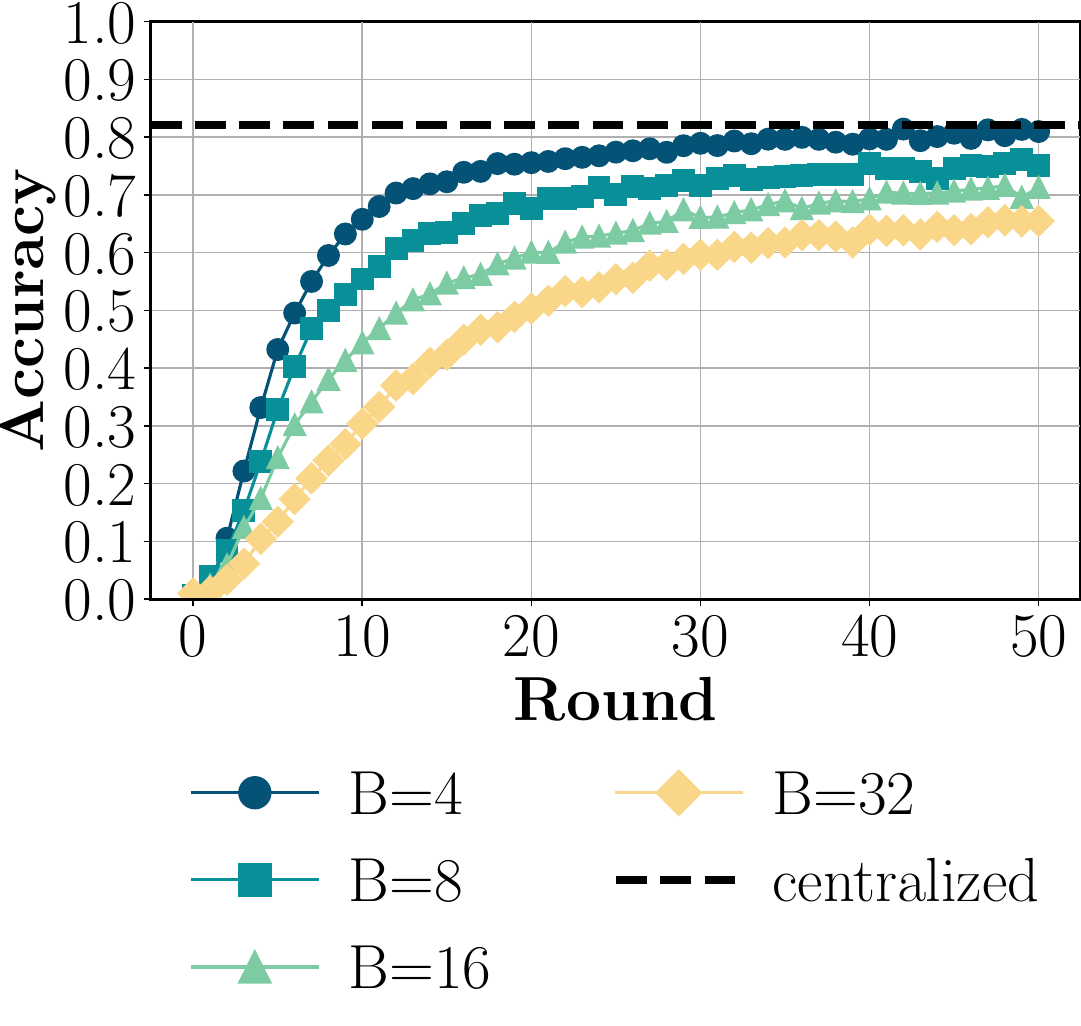}
        \caption{Video action recognition}
        \label{fig:batch_video}
    \end{subfigure}
    \caption{Results for varying client-local training batch size. `B' denotes the batch size.}
    \label{fig:batch}
\end{figure*}

\subsection{Impact of Batching}
Figure~\ref{fig:batch} shows the detailed results, which we analyze below.

\heading{Smaller batch sizes improve short-term accuracy.}
Smaller batch sizes allow better exploration of individual examples, leading to higher short-term accuracy.
This finding is best reflected in audio (Figure~\ref{fig:batch_audio}) and video (Figure~\ref{fig:batch_video}) tasks: the fast-learnt accuracy is 94.3\% (audio) and 63.2\% (video) with batch size = 4 vs. 84\% (audio) and 26.8\% (video) with batch size = 32.
We can generalize the benefits of smaller batch sizes to the other tasks (Figures~\ref{fig:batch_text} and~\ref{fig:batch_image}) as well.
This implies that to obtain a model with higher accuracy given limited training time, a small batch size is preferred.

\heading{Non-linear relationship between batch size and accuracy.}
While smaller batch sizes are generally more beneficial, further reducing the size beyond a certain point does not yield higher accuracy.
For instance, in image classification, batch size 50 provides 49.5\% and 68.9\% fast-learnt accuracy and best-learnt accuracy, which are both higher than the respective accuracy with batch size 10 (39.5\% and 55\%).
Hence, batch size should be considered as a hyperparameter that needs to be tuned for specific tasks.

\heading{Large batch sizes work better with more training rounds.}
When we look at the long-term accuracy metric, the difference between different batch sizes become smaller.
Specifically, when we compare the largest batch size to the optimal batch size for each task regarding the best-learnt accuracy (0.7\%, 5.4\%, 10.1\%, 36.4\% as the accuracy difference for text, image, audio, and video tasks), the gap is smaller compared to fast-learnt accuracy (0.8\%, 5\%, 0.5\%, 15.8\% in the same task order).
This result shows a positive message for large batch sizes, which can better utilize GPU hardware resources: higher model accuracy can be achieved with more training rounds.



\heading{Compared to centralized learning.}
FL incurs minimal accuracy overhead with the optimal batch size as Figure~\ref{fig:batch} shows---the accuracy degradation of the text and image tasks (Figures~\ref{fig:batch_text} and~\ref{fig:batch_image}) are primarily caused by non-IID data distribution.
Different batch sizes have negligible impact on model accuracy for audio classification (the best-learnt accuracy is the same as centralized learning accuracy).
For video action recognition, a large batch size (e.g., 32) can significantly slow down model convergence: after 50 rounds, the gap between FL and centralized learning remains large (65.6\% vs. 82.1\%).
This impact can be mitigated with more training rounds.

\subsection{Impact of Training Epochs}
\begin{figure*}
    \centering
    \begin{subfigure}{0.245\textwidth}
        \centering
        \includegraphics[width=\linewidth]{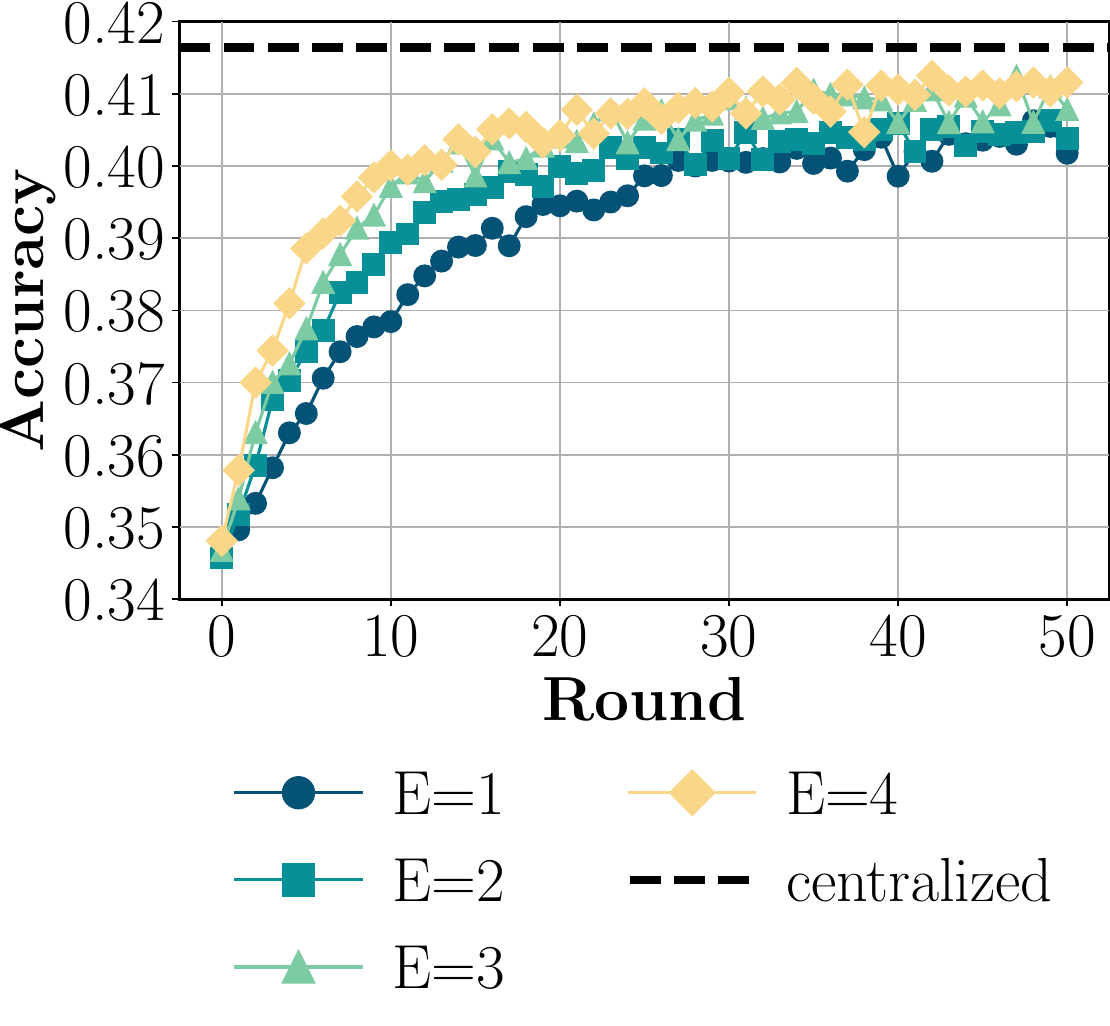}
        \caption{Next-word prediction}
        \label{fig:text_epoch}
    \end{subfigure}
    \begin{subfigure}{0.24\textwidth}
        \centering
        \includegraphics[width=\linewidth]{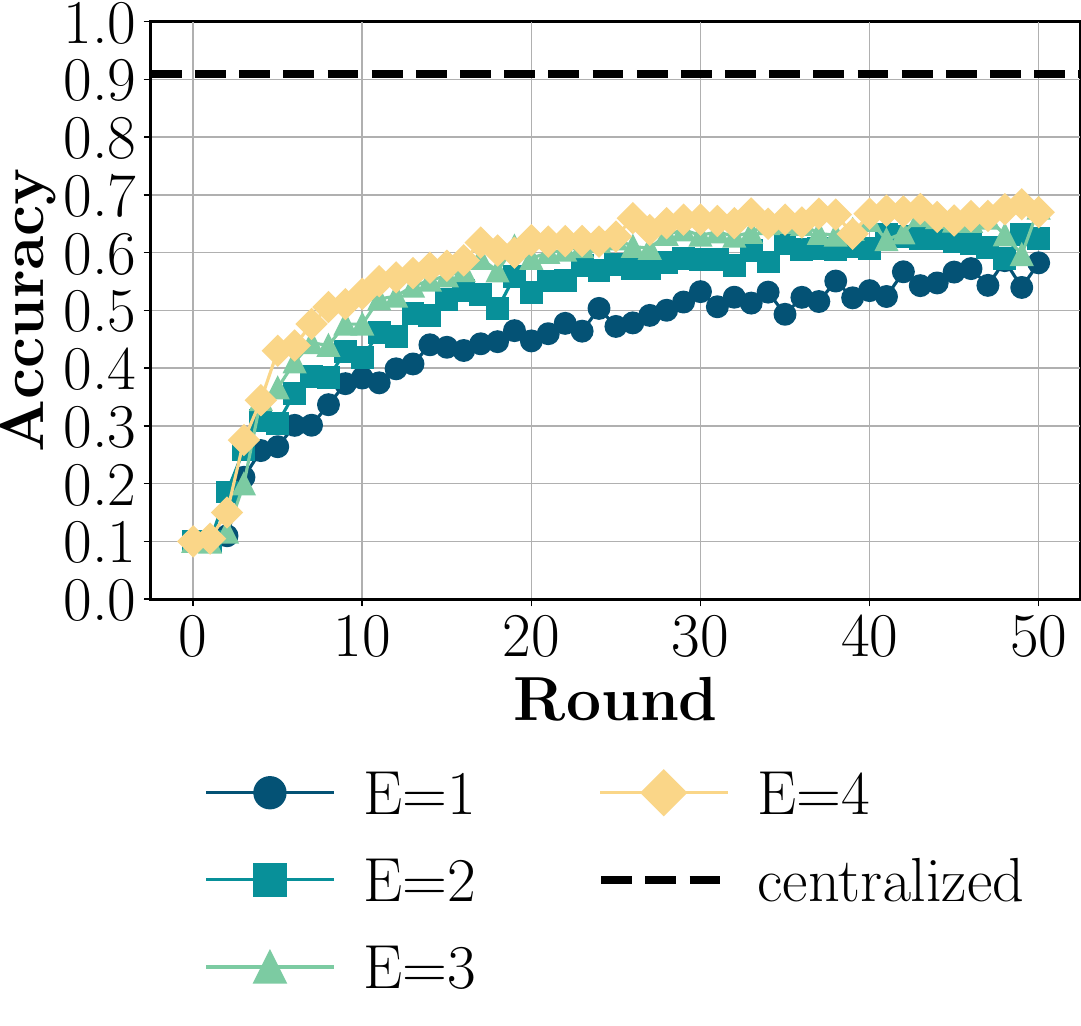}
        \caption{Image classification}
        \label{fig:image_epoch}
    \end{subfigure}
    \begin{subfigure}{0.24\textwidth}
        \centering
        \includegraphics[width=\linewidth]{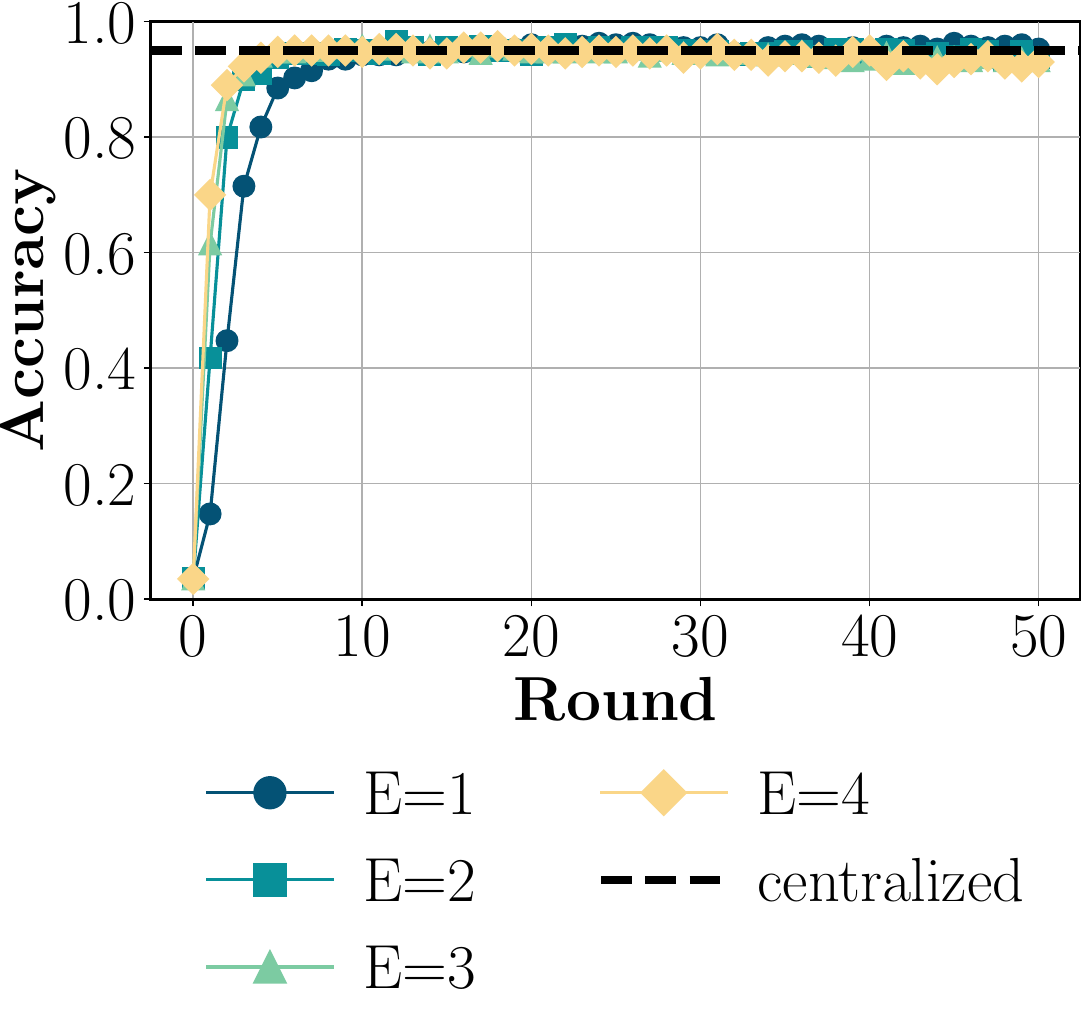}
        \caption{Audio classification}
        \label{fig:audio_epoch}
    \end{subfigure}
    \begin{subfigure}{0.24\textwidth}
        \centering
        \includegraphics[width=\linewidth]{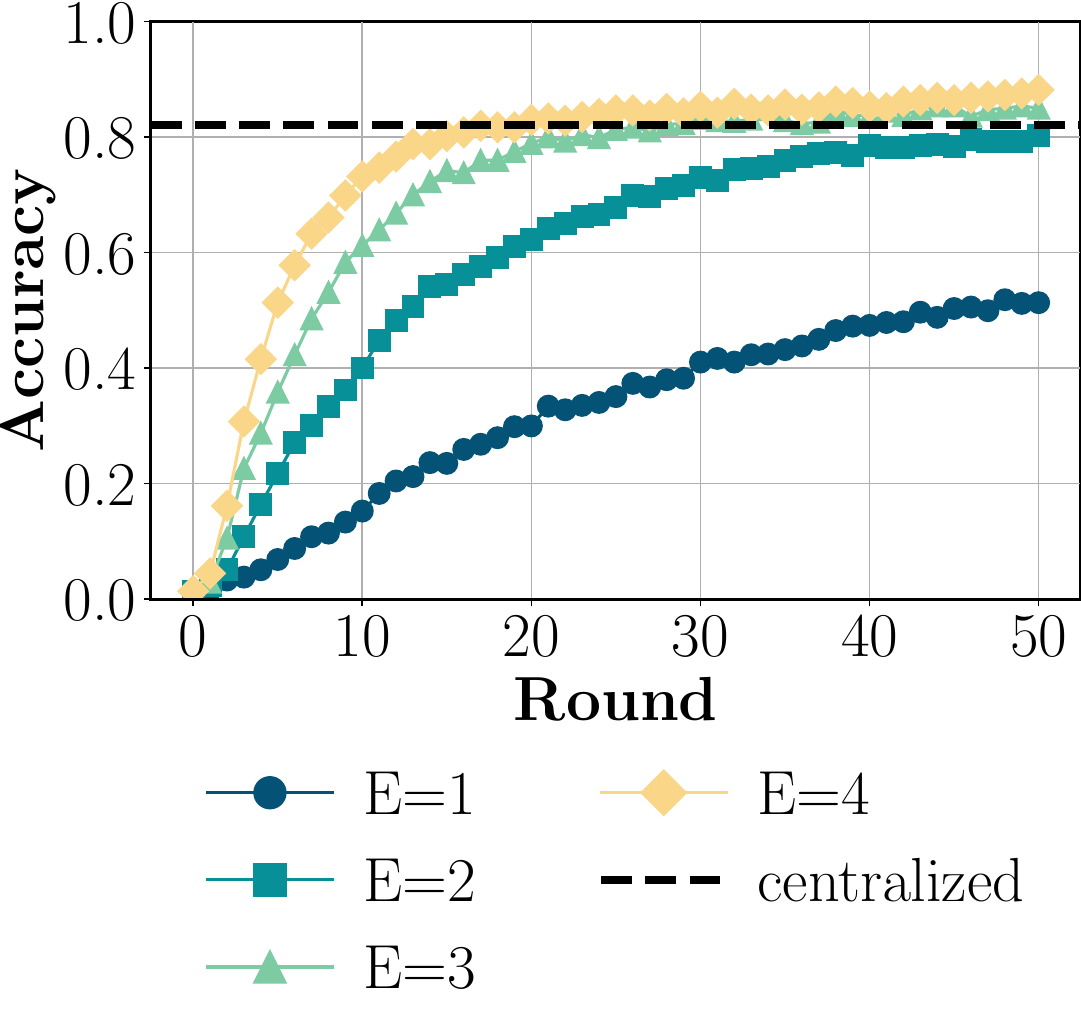}
        \caption{Video action recognition}
        \label{fig:video_epoch}
    \end{subfigure}
    \caption{Results for varying client-local training length per FL round. `E' denotes the number of local epochs.}
    \label{fig:epoch}
\end{figure*}
Figure~\ref{fig:epoch} shows the detailed results, which we analyze below.

\heading{Increasing epochs per FL round creates a similar effect of increasing FL rounds.}
We found more local training epochs reduce the FL rounds needed to reach a certain accuracy but increase the time per round.
With more epochs per round, the fast-learnt (measured by rounds) is expected to be significantly improved.
Indeed, in the four tasks (Figure~\ref{fig:epoch}),
the fast-learnt accuracy increases from 37.8\%, 37.3\%, 93.5\%, and 13.3\% to 39.8\%, 51.1\%, 95\%, and 69.9\%, respectively, when increasing the epochs per round from 1 to 4.
Yet, as training rounds increase, the gap between these two configurations narrows for all tasks.
Seemingly trivial for accuracy, 
the number of epochs per round enhances FL performance by addressing client heterogeneity.
As some clients are more resourceful and powerful than others, we may adapt local training epochs based on client capacities to mitigate stragglers, as explored in previous work~\cite{PARK202213}.



\heading{Compared to centralized learning.}
As fewer epochs per round increase rounds-to-accuracy, the best-learnt accuracy within 50 epochs can be significantly lower than that of centralized learning.
For text, image, and video predictions, the accuracy difference between centralized learning and FL (with 1 epoch per round) is 1\%, 32.3\%, and 30.3\%, respectively.
However, this gap is expected to close given more training rounds.
Audio classification is the task least sensitive to FL.
Although lower epochs indeed slow down convergence, the model can ramp up quickly with only a few more rounds and achieve the same accuracy as centralized learning.

\section{Global Federation Strategies}
\label{sec:global}
This section examines the impact of different FL algorithms on the global training process.
Specifically, we evaluate three popular FL algorithms: FedAvg~\cite{fl_origin}, FedAdam, and FedYogi~\cite{fedadam}.

FedAvg is the foundational FL algorithm, improving communication efficiency by averaging model updates from all clients. While effective for IID data, it struggles with unbalanced data distributions.
FedAdam adapts the Adam optimizer for adaptive learning rate adjustment based on the momentum of gradients and variance estimates to improve model convergence.
FedYogi adapts the Yogi optimizer, which is similar to FedAdam but handles variance updates more conservatively. 

Figure~\ref{fig:global} shows the results of our investigation.

\subsection{Impact of Federation}


\begin{figure*}
    \centering
    \begin{subfigure}{0.25\textwidth}
        \centering
        \raisebox{0.24cm}{\includegraphics[width=\linewidth]{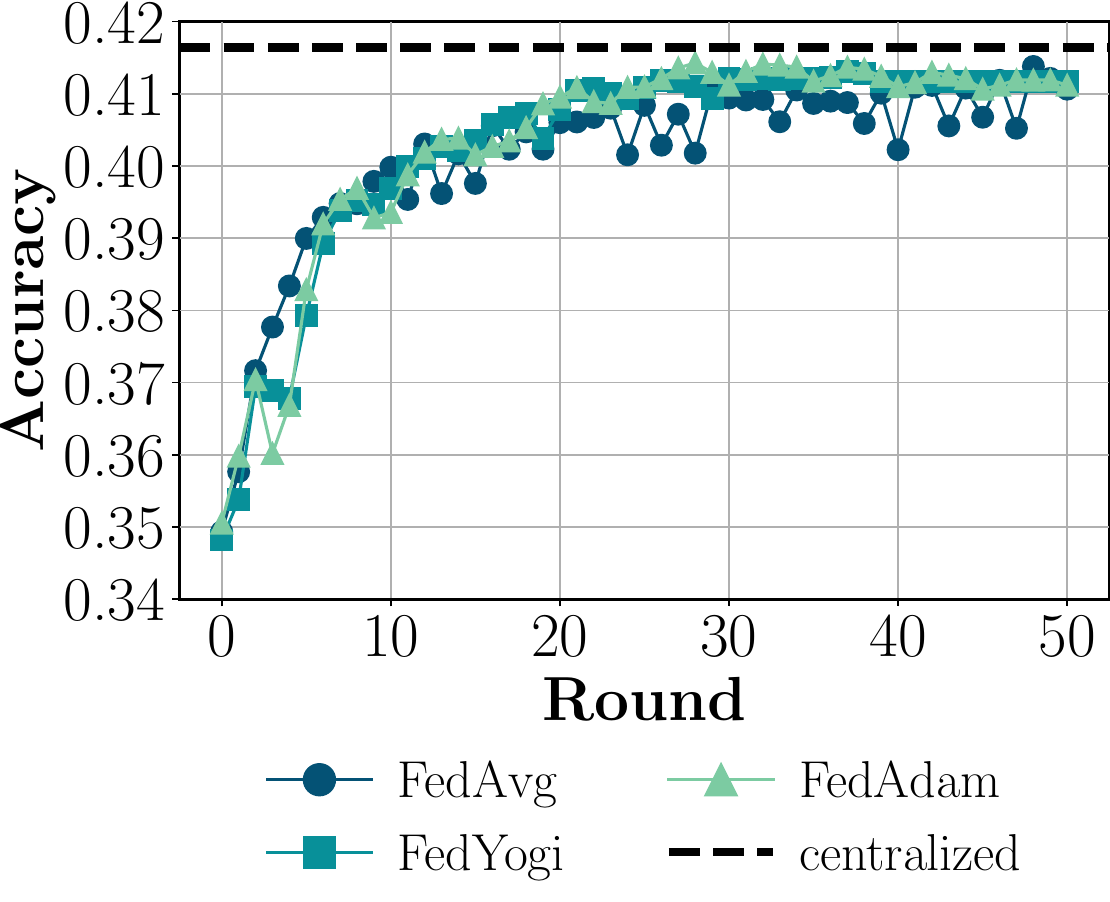}}
        \caption{Next-word prediction}
        \label{fig:global_text}
    \end{subfigure}
    \begin{subfigure}{0.25\textwidth}
        \centering
        \includegraphics[width=\linewidth, height=3.85cm]{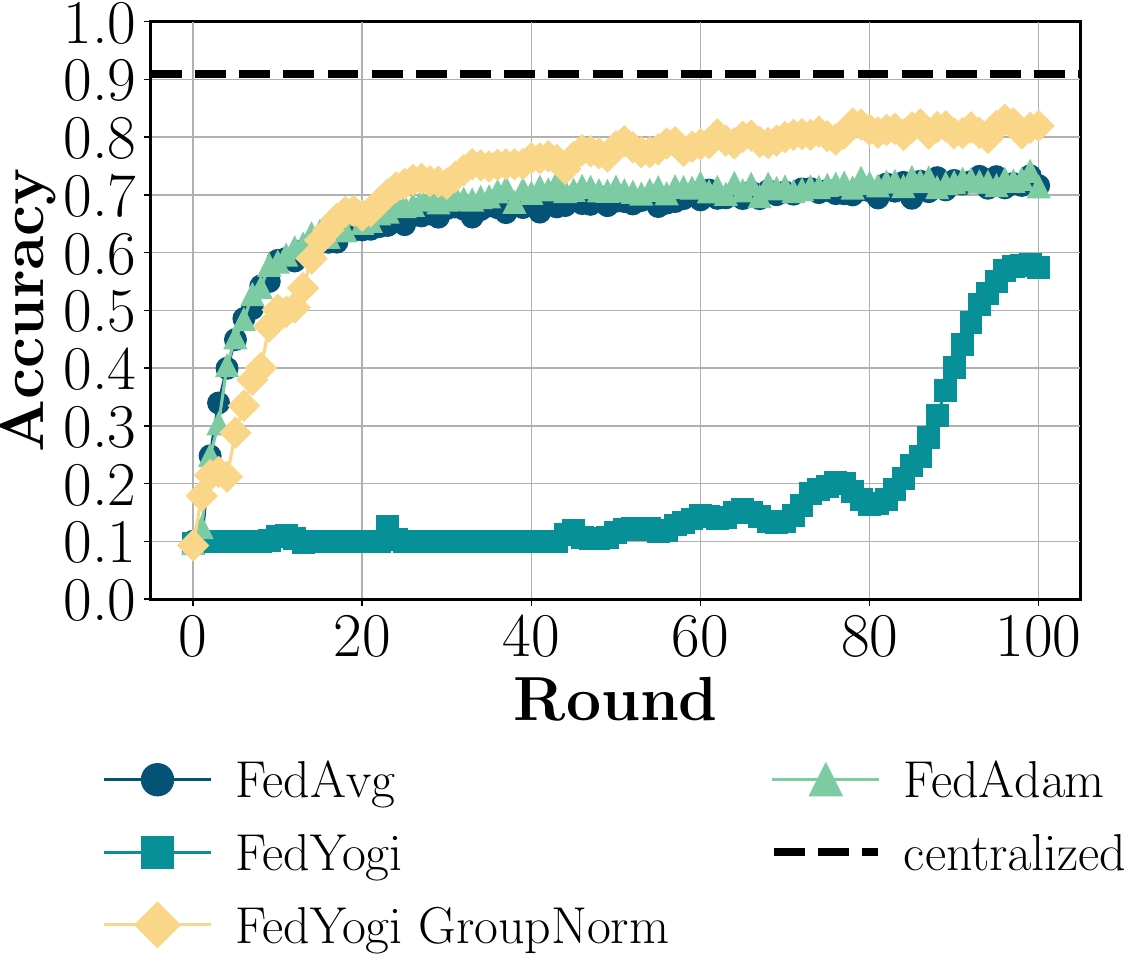}
        \caption{Image classification}
        \label{fig:global_image}
    \end{subfigure}
    \begin{subfigure}{0.24\textwidth}
        \centering
        \raisebox{0.30cm}{\includegraphics[width=\linewidth]{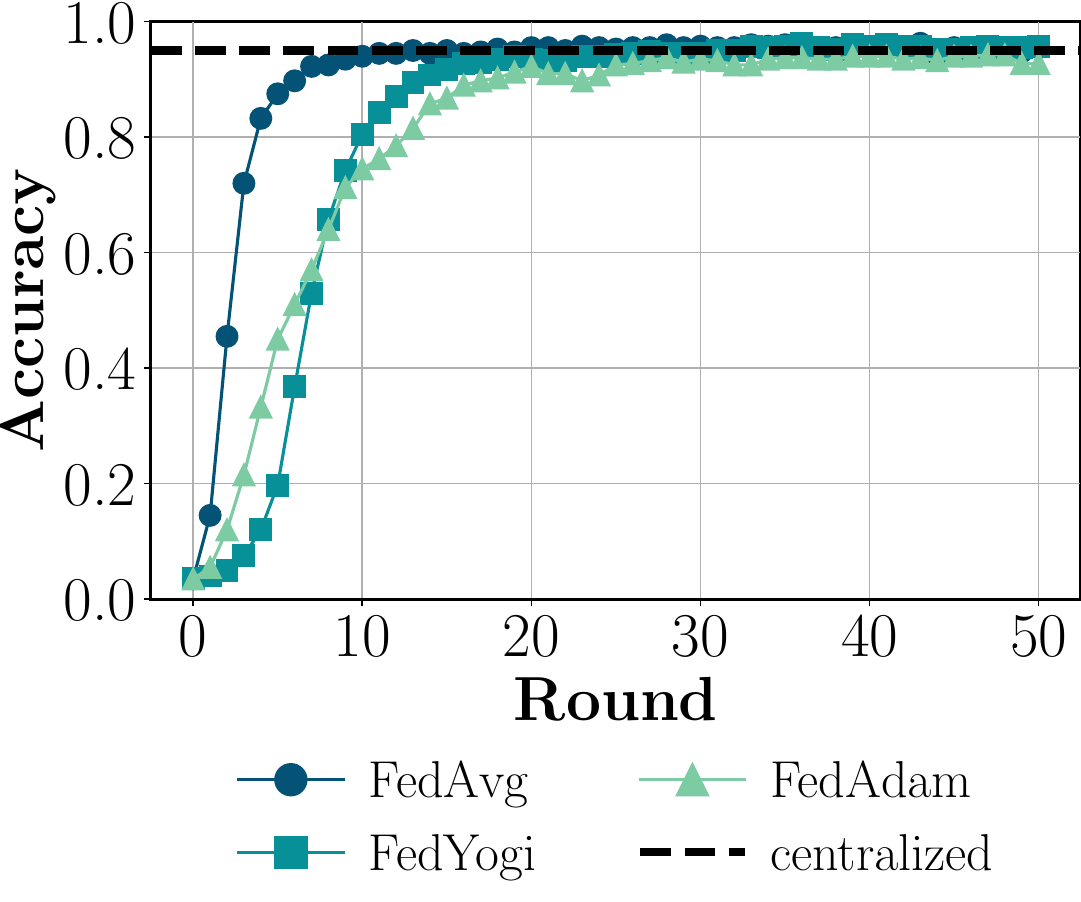}}
        \caption{Audio classification}
        \label{fig:global_audio}
    \end{subfigure}
    \begin{subfigure}{0.24\textwidth}
        \centering
        \raisebox{0.30cm}{\includegraphics[width=\linewidth]{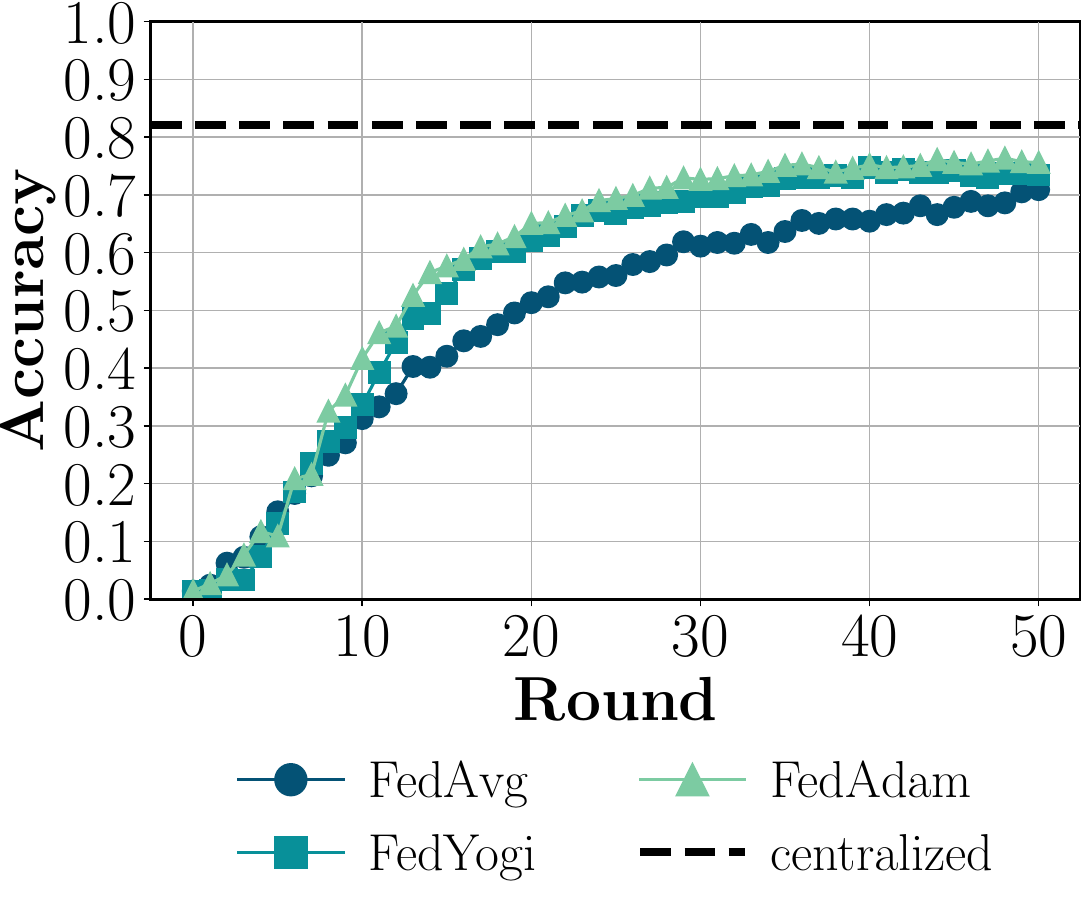}}
        \caption{Video action recognition}
        \label{fig:global_video}
    \end{subfigure}
    \caption{Results for different global federation strategies.}
    \label{fig:global}
\end{figure*}

\heading{Algorithm performance varies across tasks.}
The effectiveness of each algorithm varies significantly depending on the task. 
For text (Figure~\ref{fig:global_text}), all three algorithms perform similarly, with FedAdam slightly better than the others (best-learnt accuracy 41.4\% compared to 41.4\% for FedAvg and 41.3\% for FedYogi).
In image classification (Figure~\ref{fig:global_image}) , FedYogi significantly outperforms the other algorithms, achieving a best-learnt accuracy of 83\% compared to 73.3\% for FedAvg and 74.1\% for FedAdam. 
For audio classification (Figures~\ref{fig:global_audio}), FedAvg and FedYogi achieve the same best-learnt accuracy of 96.3\%, slightly outperforming FedAdam (94.3\%). 
In video action recognition (Figure~\ref{fig:global_video}), FedAdam shows the best performance with a best-learnt accuracy of 76.4\%, followed closely by FedYogi (74.7\%) and FedAvg (70.9\%).

\heading{FL algorithm and model architecture can be incompatible.} 
FedYogi encounters compatibility problems with certain model architectures.
For the image task, the original FedYogi (with BatchNorm) performed poorly on image classification, achieving only 57.9\% best-learnt accuracy (Figure~\ref{fig:global_image}) due to inaccurate estimation of global mean and variance from mini-batches.
Replacing BatchNorm with GroupNorm resolved this issue, dramatically improving FedYogi's performance to 83\% best-learnt accuracy. 

\heading{FedAvg is effective.} 
Despite being the simplest algorithm, FedAvg demonstrates competitive performance across all tasks.
In text prediction and audio classification, FedAvg achieves the highest fast-learnt accuracy (39.8\% and 93.5\% respectively), indicating faster initial convergence.
For image and video tasks, FedAvg's best-learnt accuracy remains competitive, within 3-6\% of the best-performing algorithm. This indicates that FedAvg remains a strong baseline.
More advanced algorithms like FedAdam and FedYogi can offer improvements in certain scenarios, e.g., when data is non-IID distributed (FedYogi in Figure~\ref{fig:global_image}). 
\section{Related Work}
\label{sec:related}

\noindent \textbf{FL Domain Applications:} FL has been explored across various domains, demonstrating its effectiveness in handling decentralized data while preserving privacy. In healthcare, FL has been applied to health data analysis, such as detecting depression~\cite{fl_app_med1}. In recommendation systems, FL has been used for personalized recommendations in mobile and IoT scenarios and leverages techniques like co-clustering and secure low-rank training for cross-domain recommendations~\cite{fl_app_rec1, fl_app_rec2, fl_app_rec3, fl_app_rec4, fl_app_rec5, fl_app_rec6}. FL also facilitates mobile crowdsensing in the context of smart cities, including energy grids, transportation services, and water distribution to smart homes~\cite{fl_app_smartcity1, fl_app_smartcity2}. In finance, FL enables secure collaboration among institutions for risk assessment and fraud detection~\cite{fl_app_fin1, fl_app_fin2}. FL has also been applied in edge computing for distributed edge devices, such as vehicles~\cite{fl_app_edge1}, and satellites~\cite{fedspace}. Recently, FL has been utilized for training large language models~\cite{fl_app_llm1, fl_app_llm2}.

\noindent \textbf{FL Related Survey:} Several surveys have covered different aspects of FL, including its basic concepts~\cite{yang2019federated}, challenges in large-scale mobile and edge environments~\cite{li2020federated, kairouz2021advances}, platforms and protocols~\cite{aledhari2020federated}, potential threats to FL~\cite{lyu2020threats}, categorization of FL systems~\cite{li2021survey}, and the impact of non-IID data in FL~\cite{li2022federated}. Despite these surveys on FL, a gap remains in evaluating the FL's impact on model accuracy. Thus,  we conduct a comprehensive experimental study that examines the effect of FL on the accuracy of  state-of-the-art machine learning models across various types of tasks.



\section{Conclusion}
\label{sec:conclusion}

This paper presents the results of an extensive set of experiments to evaluate the impact of FL, an ML paradigm that can utilize globally distributed data, on the accuracy of state-of-the-art models for various ML tasks, covering text, image, audio, and video data using a unified FL framework.
The detailed and quantitative analysis centers around crucial FL aspects: data distribution, client sampling, scale, as well as client-local and global computations.
Our findings reveal how each of these aspects incurs accuracy overhead, providing valuable information for future FL development.


\balance
\bibliographystyle{abbrv}
\bibliography{ref}

\end{document}